\documentclass[10pt,twocolumn,letterpaper]{article}

\usepackage{cvpr}              

\usepackage[dvipsnames]{xcolor}

\definecolor{cvprblue}{rgb}{0.21,0.49,0.74}
\usepackage[pagebackref,breaklinks,colorlinks,citecolor=cvprblue]{hyperref}
\usepackage{algorithm}
\usepackage{algorithmic}
\usepackage{csquotes}
\usepackage{multirow}

\def\paperID{6676} 
\def\confName{CVPR}
\def\confYear{2024}

\title{Unifying Top-down and Bottom-up Scanpath Prediction Using Transformers}

\author{Zhibo Yang\textsuperscript{1,2}, Sounak Mondal\textsuperscript{1}, Seoyoung Ahn\textsuperscript{1}, Ruoyu Xue\textsuperscript{1}, \\  Gregory Zelinsky\textsuperscript{1}, Minh Hoai\textsuperscript{1,3}, Dimitris Samaras\textsuperscript{1} \\
\textsuperscript{1}Stony Brook University~~~  \textsuperscript{2}Waymo LLC~~~
\textsuperscript{3}VinAI Research
}

\begin{document}
\maketitle

\def\mA{\mathcal{A}}
\def\mB{\mathcal{B}}
\def\mC{\mathcal{C}}
\def\mD{\mathcal{D}}
\def\mE{\mathcal{E}}
\def\mF{\mathcal{F}}
\def\mG{\mathcal{G}}
\def\mH{\mathcal{H}}
\def\mI{\mathcal{I}}
\def\mJ{\mathcal{J}}
\def\mK{\mathcal{K}}
\def\mL{\mathcal{L}}
\def\mM{\mathcal{M}}
\def\mN{\mathcal{N}}
\def\mO{\mathcal{O}}
\def\mP{\mathcal{P}}
\def\mQ{\mathcal{Q}}
\def\mR{\mathcal{R}}
\def\mS{\mathcal{S}}
\def\mT{\mathcal{T}}
\def\mU{\mathcal{U}}
\def\mV{\mathcal{V}}
\def\mW{\mathcal{W}}
\def\mX{\mathcal{X}}
\def\mY{\mathcal{Y}}
\def\mZ{\mathcal{Z}} 

\def\bbN{\mathbb{N}} 
\def\bbR{\mathbb{R}} 
\def\bbP{\mathbb{P}} 
\def\bbQ{\mathbb{Q}} 
\def\bbE{\mathbb{E}}

\def\1n{\mathbf{1}_n}
\def\0{\mathbf{0}}
\def\1{\mathbf{1}}

\def\A{{\bf A}}
\def\B{{\bf B}}
\def\C{{\bf C}}
\def\D{{\bf D}}
\def\E{{\bf E}}
\def\F{{\bf F}}
\def\G{{\bf G}}
\def\H{{\bf H}}
\def\I{{\bf I}}
\def\J{{\bf J}}
\def\K{{\bf K}}
\def\L{{\bf L}}
\def\M{{\bf M}}
\def\N{{\bf N}}
\def\O{{\bf O}}
\def\P{{\bf P}}
\def\Q{{\bf Q}}
\def\R{{\bf R}}
\def\S{{\bf S}}
\def\T{{\bf T}}
\def\U{{\bf U}}
\def\V{{\bf V}}
\def\W{{\bf W}}
\def\X{{\bf X}}
\def\Y{{\bf Y}}
\def\Z{{\bf Z}}

\def\a{{\bf a}}
\def\b{{\bf b}}
\def\c{{\bf c}}
\def\d{{\bf d}}
\def\e{{\bf e}}
\def\f{{\bf f}}
\def\g{{\bf g}}
\def\h{{\bf h}}
\def\i{{\bf i}}
\def\j{{\bf j}}
\def\k{{\bf k}}
\def\l{{\bf l}}
\def\m{{\bf m}}
\def\n{{\bf n}}
\def\o{{\bf o}}
\def\p{{\bf p}}
\def\q{{\bf q}}
\def\r{{\bf r}}
\def\s{{\bf s}}
\def\t{{\bf t}}
\def\u{{\bf u}}
\def\v{{\bf v}}
\def\w{{\bf w}}
\def\x{{\bf x}}
\def\y{{\bf y}}
\def\z{{\bf z}}

\def\balpha{\mbox{\boldmath{$\alpha$}}}
\def\bbeta{\mbox{\boldmath{$\beta$}}}
\def\bdelta{\mbox{\boldmath{$\delta$}}}
\def\bgamma{\mbox{\boldmath{$\gamma$}}}
\def\blambda{\mbox{\boldmath{$\lambda$}}}
\def\bsigma{\mbox{\boldmath{$\sigma$}}}
\def\btheta{\mbox{\boldmath{$\theta$}}}
\def\bomega{\mbox{\boldmath{$\omega$}}}
\def\bxi{\mbox{\boldmath{$\xi$}}}
\def\bnu{\mbox{\boldmath{$\nu$}}}                                  
\def\bphi{\mbox{\boldmath{$\phi$}}}
\def\bmu{\mbox{\boldmath{$\mu$}}}

\def\bDelta{\mbox{\boldmath{$\Delta$}}}
\def\bOmega{\mbox{\boldmath{$\Omega$}}}
\def\bPhi{\mbox{\boldmath{$\Phi$}}}
\def\bLambda{\mbox{\boldmath{$\Lambda$}}}
\def\bSigma{\mbox{\boldmath{$\Sigma$}}}
\def\bGamma{\mbox{\boldmath{$\Gamma$}}}
                                  
\newcommand{\myprob}[1]{\mathop{\mathbb{P}}_{#1}}

\newcommand{\myexp}[1]{\mathop{\mathbb{E}}_{#1}}

\newcommand{\mydelta}[1]{1_{#1}}

\newcommand{\myminimum}[1]{\mathop{\textrm{minimum}}_{#1}}
\newcommand{\mymaximum}[1]{\mathop{\textrm{maximum}}_{#1}}    
\newcommand{\mymin}[1]{\mathop{\textrm{minimize}}_{#1}}
\newcommand{\mymax}[1]{\mathop{\textrm{maximize}}_{#1}}
\newcommand{\mymins}[1]{\mathop{\textrm{min.}}_{#1}}
\newcommand{\mymaxs}[1]{\mathop{\textrm{max.}}_{#1}}  
\newcommand{\myargmin}[1]{\mathop{\textrm{argmin}}_{#1}} 
\newcommand{\myargmax}[1]{\mathop{\textrm{argmax}}_{#1}} 
\newcommand{\myst}{\textrm{s.t. }}

\newcommand{\denselist}{\itemsep -1pt}
\newcommand{\sparselist}{\itemsep 1pt}

\definecolor{pink}{rgb}{0.9,0.5,0.5}
\definecolor{purple}{rgb}{0.5, 0.4, 0.8}   
\definecolor{gray}{rgb}{0.3, 0.3, 0.3}
\definecolor{mygreen}{rgb}{0.2, 0.6, 0.2}

\newcommand{\cyan}[1]{\textcolor{cyan}{#1}}
\newcommand{\blue}[1]{\textcolor{blue}{#1}}
\newcommand{\magenta}[1]{\textcolor{magenta}{#1}}
\newcommand{\pink}[1]{\textcolor{pink}{#1}}
\newcommand{\green}[1]{\textcolor{green}{#1}} 
\newcommand{\gray}[1]{\textcolor{gray}{#1}}    
\newcommand{\mygreen}[1]{\textcolor{mygreen}{#1}}    
\newcommand{\purple}[1]{\textcolor{purple}{#1}}       

\definecolor{greena}{rgb}{0.4, 0.5, 0.1}
\newcommand{\greena}[1]{\textcolor{greena}{#1}}

\definecolor{bluea}{rgb}{0, 0.4, 0.6}
\newcommand{\bluea}[1]{\textcolor{bluea}{#1}}
\definecolor{reda}{rgb}{0.6, 0.2, 0.1}
\newcommand{\reda}[1]{\textcolor{reda}{#1}}

\def\changemargin#1#2{\list{}{\rightmargin#2\leftmargin#1}\item[]}
\let\endchangemargin=\endlist
                                               
\newcommand{\cm}[1]{}

\newcommand{\mhoai}[1]{{\color{magenta}\textbf{[MH: #1]}}}
\newcommand{\ruoyux}[1]{{\color{purple}\textbf{[RX: #1]}}}

\newcommand{\mtodo}[1]{{\color{red}$\blacksquare$\textbf{[TODO: #1]}}}
\newcommand{\myheading}[1]{\vspace{1ex}\noindent \textbf{#1}}
\newcommand{\htimesw}[2]{\mbox{$#1$$\times$$#2$}}

\newcommand{\young}[1]{{\color{blue}$\blacksquare$\textbf{Alternative}: #1}}


\newif\ifshowsolution
\showsolutiontrue

\ifshowsolution  
\newcommand{\Comment}[1]{\paragraph{\bf $\bigstar $ COMMENT:} {\sf #1} \bigskip}
\newcommand{\Solution}[2]{\paragraph{\bf $\bigstar $ SOLUTION:} {\sf #2} }
\newcommand{\Mistake}[2]{\paragraph{\bf $\blacksquare$ COMMON MISTAKE #1:} {\sf #2} \bigskip}
\else
\newcommand{\Solution}[2]{\vspace{#1}}
\fi

\newcommand{\truefalse}{
\begin{enumerate}
	\item True
	\item False
\end{enumerate}
}

\newcommand{\yesno}{
\begin{enumerate}
	\item Yes
	\item No
\end{enumerate}
}

\newcommand{\Sref}[1]{Sec.~\ref{#1}}
\newcommand{\Eref}[1]{Eq.~(\ref{#1})}
\newcommand{\Fref}[1]{Fig.~\ref{#1}}
\newcommand{\Tref}[1]{Table~\ref{#1}}
\begin{abstract}
Most models of visual attention aim at predicting either top-down or bottom-up control, as studied using different visual search and free-viewing tasks. In this paper we propose the Human Attention Transformer (HAT), a single model that predicts both forms of attention control. HAT uses a novel transformer-based architecture and a simplified foveated retina that collectively create a spatio-temporal awareness akin to the dynamic visual working memory of humans. HAT not only establishes a new state-of-the-art in predicting the scanpath of fixations made during target-present and target-absent visual search and ``taskless'' free viewing, but also makes human gaze behavior interpretable. Unlike previous methods that rely on a coarse grid of fixation cells and experience information loss due to fixation discretization, HAT features a sequential dense prediction architecture and outputs a dense heatmap for each fixation, thus avoiding discretizing fixations. HAT sets a new standard in computational attention, which emphasizes effectiveness, generality, and interpretability. HAT's demonstrated scope and applicability will likely inspire the development of new attention models that can better predict human behavior in various attention-demanding scenarios. Code is available at \href{https://github.com/cvlab-stonybrook/HAT}{https://github.com/cvlab-stonybrook/HAT}.
\end{abstract}    
\section{Introduction}
\label{sec:intro}
Attention, a cognitive process that allows humans to selectively allocate their limited cognitive resources to specific regions of the visual world, plays a crucial role in human perception system. Understanding and predicting human (visual) attention will enable numerous applications such as assistive technologies that can anticipate a person's needs and intents, perceptions system that can prioritize processing regions of human interest and enhancing the accuracy and speed of various visual tasks (e.g., object detection), and image/video compression that allocates more resources to encoding and transmitting high-attention regions, optimizing the use of bandwidth.

Human attention control can take two broad forms. One is bottom-up, meaning that attention saliency signals are computed from the visual input and used to prioritize shifts of attention. The same visual input should therefore lead to the same shifts of bottom-up attention. The second type of attention is top-down, meaning that a task or goal is used to control attention. Given a kitchen scene, very different fixations are observed depending on whether a person is searching for a clock or a microwave oven \cite{zelinsky2021predicting}. These two types of attention control spawned two separate literatures on gaze fixation prediction (the accepted measure of attention), one where studies use a free-viewing task to study questions of bottom-up attention and the other using a goal-directed task (typically, visual search) to study top-down attention control.
Consequently, most models have been designed to address {\it either} bottom-up {\it or} top-down attention, not both. {\it Can a single model architecture predict both bottom-up and top-down attention control?}

\begin{figure}[t]
  \centering
  \includegraphics[width=1.0\linewidth]{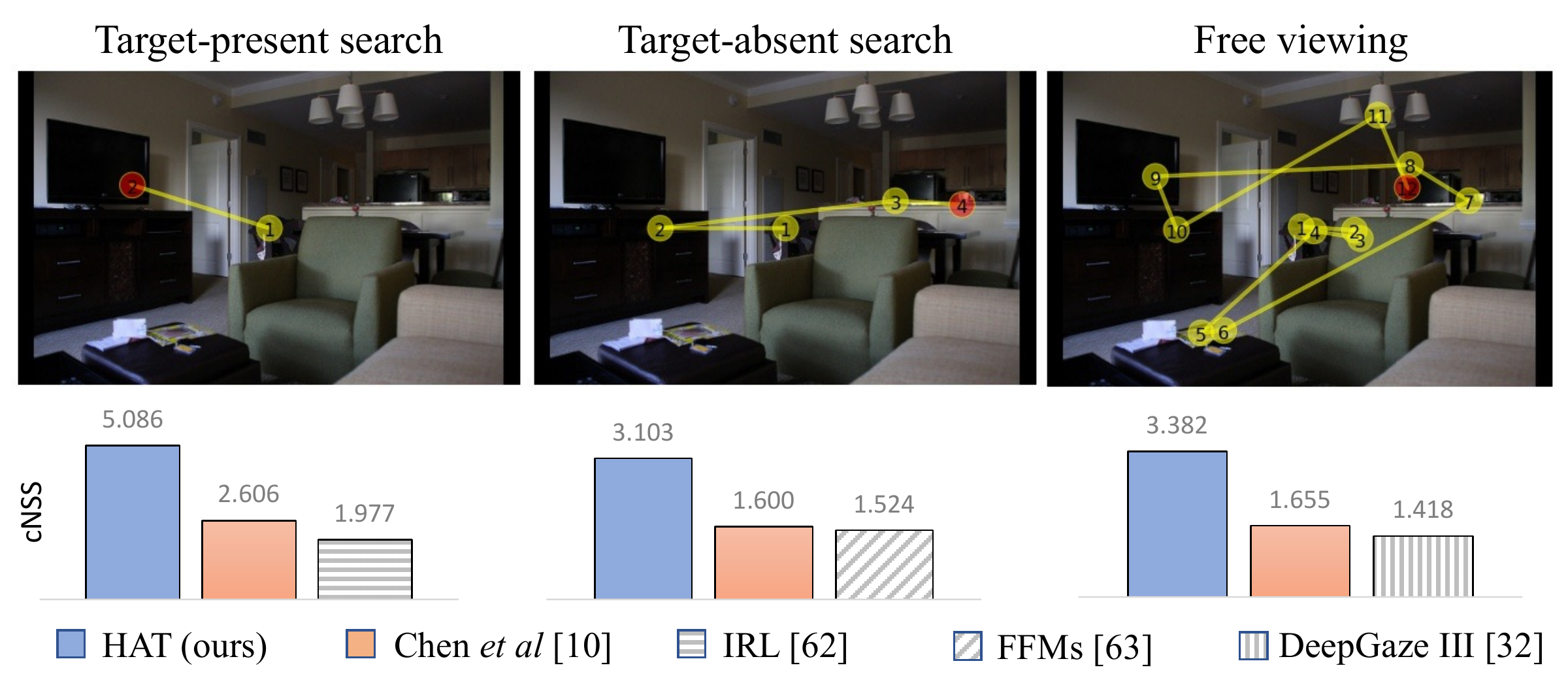}
  \caption{Given an image, the proposed {\bf HAT} is able to predict scanpaths under three settings target-present search for TV; target-absent scanpath for sink; and free viewing. Importantly, HAT outperforms previous state-of-the-art scanpath prediction methods on multiple datasets across three settings: target-present, target-absent visual search and free viewing, that were studied separately.}
  \label{fig:teaser}
\end{figure}

Our answer to this question is HAT, a Human Attention Transformer that generally predicts scanpaths of fixations, meaning that it can be applied to both top-down visual search and bottom-up free viewing tasks (\Cref{fig:teaser}). Devising a unified model architecture capable of predicting both bottom-up and top-down attention control is nontrivial: 1) predicting human fixation scanpaths requires the model to have a spatio-temporal understanding of the fixated image contents and their relationship to the external goals; and 2) predicting top-down and bottom-up attention requies the model to capture both low-level features and high-level semantics of the input image. 
HAT addresses these issues by using a novel transformer-based design and a simplified foveated retina. Together, these components forge a novel paradigm, constituting a form of {\it dynamically-updating visual working memory}.
Traditional approaches have leaned on recurrent neural networks (RNNs) to uphold a dynamically updated hidden vector conveying information across fixations \cite{chen2021predicting,sun2019visual,zelinsky2019benchmarking,assens2018pathgan}.
Alternatively, simulations of a foveated retina have combined multi-resolution information at pixel \cite{zelinsky2019benchmarking}, feature \cite{yang2022target}, or semantic levels \cite{yang2020predicting}.
However, these methods present drawbacks: RNNs sacrifice interpretability, while multi-resolution simulations fall short in capturing crucial temporal and spatial information integral for scanpath prediction.

In addressing these challenges, we leverage a computational attention mechanism \cite{vaswani2017attention} to dynamically assimilate spatial, temporal, and visual information acquired at each fixation into working memory~\cite{oberauer2019working, olivers2020attention}. This empowers HAT to discern a set of task-specific attention weights for amalgamating information from working memory and forecasting human attention control. 
This innovative mechanism sheds light on the intricate relationship between human attention and working memory~\cite{desimone1995neural, gazzaley2012top}, rendering HAT not only cognitively plausible but also ensuring the interpretability of its predictions. Furthermore, in contrast to prior methods \cite{yang2020predicting,yang2022target,chen2021predicting}, HAT treats scanpath prediction as a sequence of dense prediction tasks with per-pixel supervision, successfully avoiding the need for discretizing fixations. This enhances the method's efficacy, particularly in scenarios involving high-resolution imagery.

To demonstrate HAT's generality, we predict scanpaths
under three settings, target-present (TP) and target-absent (TA) visual search, and free-viewing (FV), covering both top-down and bottom-up attention. 
In  previous work predicting search scanpaths \cite{yang2020predicting,yang2022target,chen2021predicting}, separate models were trained for the TP and TA settings. HAT is a single model establishing new SOTA in both TP and TA search-scanpath prediction. When trained with FV scanpaths, HAT also achieves top performance relative to baselines. HAT advances SOTA in cNSS by 95\%, 94\% and 104\% under the TP, TA and FV settings on the COCO-Search18 dataset \cite{chen2021coco} and the COCO-FreeView dataset \cite{Chen_2022_CVPR}, respectively.

Our contributions can be summarized as follows:
\begin{enumerate}
    \item We propose HAT, a novel transformer architecture integrating visual information at two different eccentricities (approximating a foveated retina) to predict the spatial and temporal allocation of human attention.

    \item We formulate scanpath prediction as a sequential dense prediction task without fixation discretization, making HAT applicable to high-resolution input.
    
    \item The HAT architecture can be broadly applied to different attention control tasks, evidenced by the SOTA scanpath predictions in both visual search and free-viewing tasks. 
    
    \item HAT's attention predictions offer high interpretability, making it useful for studying gaze behavior.
\end{enumerate}
\begin{figure*}[t]
  \centering
  \includegraphics[width=1.0\linewidth]{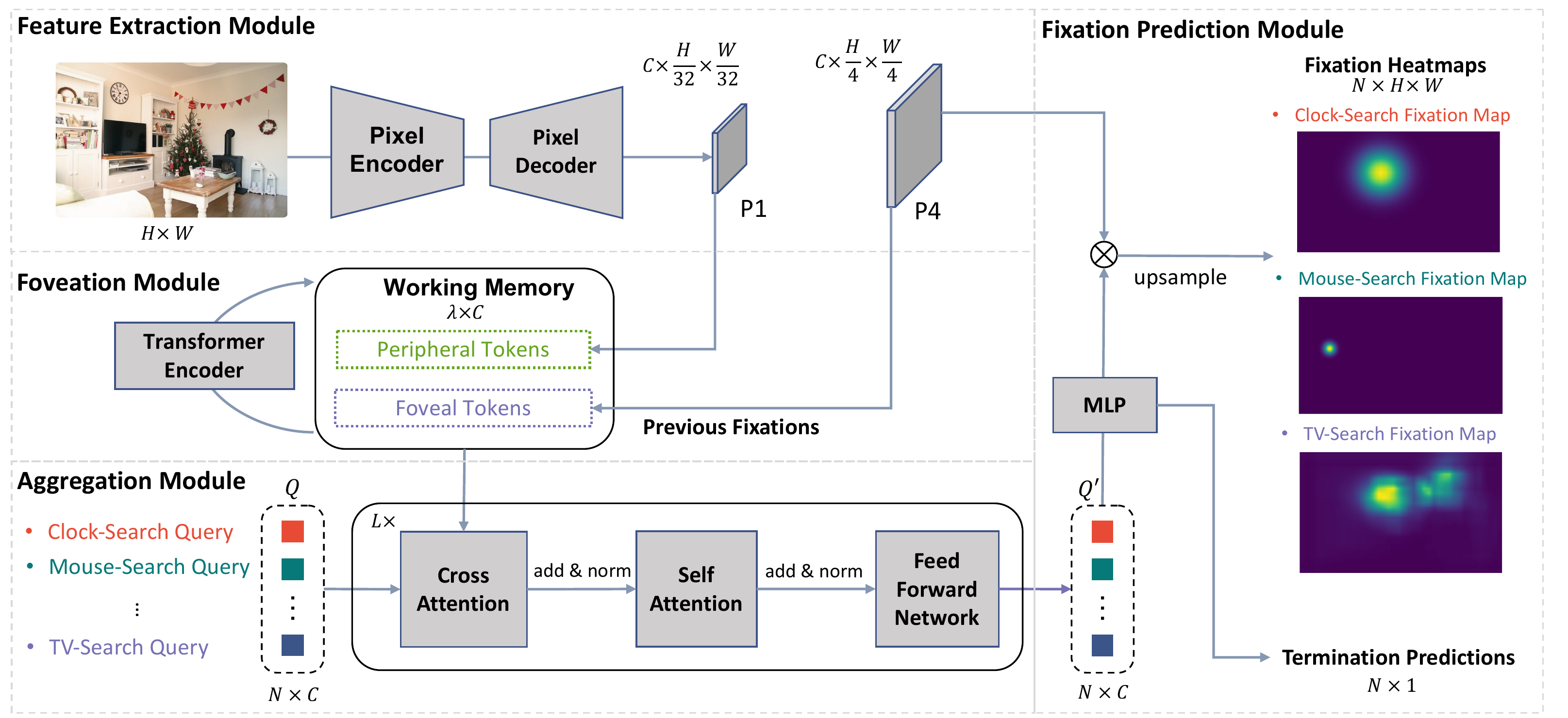}
  \caption{{\bf HAT overview.} We use encoder-decoder CNNs to extract two sets of feature maps $P_1$ and $P_4$ of different spatial resolutions. A working memory with a capacity of $\lambda$ tokens is constructed by combining all feature vectors from $P_1$ with the feature vectors of $P_4$ at previously fixated locations, representing information extracted from the periphery and central fovea. A transformer encoder is used to dynamically update the working memory at every new fixation. Then, HAT produces $N$ per-task queries of dimension $C$ (e.g., clock search and mouse search), with each learning to aggregates task-specific information from the shared working memory for predicting the fixations for its own task. Finally, the updated queries are convolved with $P_4$ to yield the fixation heatmaps after a MLP layer, and projected to the termination probabilities in parallel. Note, although this figure depicts visual search, the framework also applies for free viewing.}
  \label{fig:overview}
\end{figure*}
\section{Related Work}
\label{sec:relateWork}
\textbf{Saliency prediction.}
Predicting and understanding human gaze control has been a topic of interest for decades in psychology~\cite{yarbus1967eye, findlay2001visual, zelinsky2008theory, wolfe2017five}, but it has only recently attracted the researcher's attention in computer vision. In particular, Itti's seminal work \cite{itti2000saliency} on the saliency model has triggered a lot of interest on human attention modeling in computer vision community and facilitated many other studies identifying and modeling the salient visual features of an image (i.e., saliency prediction) ~\cite{borji2013state,kummerer2017understanding, kruthiventi2017deepfix,huang2015salicon,kummerer2014deep, cornia2018predicting,jiang2015salicon,jetley2016end, borji2015salient, masciocchi2009everyone, berg2009free,wang2017deep,wang2019revisiting}. However, the scope of these work is often narrowly focused on predicting human natural eye-movements without a specific visual task (i.e., free-viewing), ignoring another important form of attention control, goal-directed attention. Moreover, saliency models only model the spatial distribution of fixations and do not predict the temporal order between fixations. Scanpath prediction is more challenging problem because it requires predicting not only \textit{where} a fixation will be, but also \textit{when} it will be there.
\\
\textbf{Scanpath prediction.}
Many existing scanpath prediction deep neural networks (DNN) focus on predicting the free-viewing scanpaths \cite{assens2017saltinet,assens2018pathgan,sun2019visual,kummerer2022deepgaze}, primarily due to their close connection to saliency modeling. However, these models are inherently constrained in their ability to capture the full spectrum of human attention control, particularly goal-directed attention---a fundamental cognitive process that underlies various everyday visual tasks such as navigation and motor control. 
Although goal-directed human attention has been studied for decades~\cite{yarbus1967eye, land2009looking,wolfe1998visual} in cognitive science (mainly in the context of visual search~\cite{najemnik2005optimal, torralba2006contextual, zelinsky2008theory}), the development of DNNs for goal-directed scanpath prediction lags behind those designed for free-viewing tasks, partly due to the lack of data.
To tackle this problem, \citet{chen2021coco} created the first large-scale goal-directed gaze dataset with 18 search targets, COCO-Search18.
In \cite{yang2020predicting}, an inverse reinforcement learning model showed superior performance on COCO-Search18 in predicting TP scanpaths. Later, \citet{chen2021predicting} showed that a reinforcement learning model directly optimized on the scanpath similarity metric can predict VQA scanpaths, as well as on TP search scanpaths. \citet{yang2022target,rashidi2020optimal} also proposed a more generalized scanpath prediction model that can be applied to both target-present and target-absent visual search scanpaths. 
Most recently, a transformer-based scanpath prediction model, Gazeformer \cite{mondal2023gazeformer}, further advanced the TP search scanpath prediction performance on COCO-Search18. However, none of these work have demonstrated the generalizability to all three settings (i.e., TP, TA and FV). In this work, we design a generic scanpath model that generalizes to both free-viewing and visual search tasks. 
\\
\textbf{Scanpath Transformers.}
The transformative game-changing impact of Transformers \cite{vaswani2017attention} has been widely recognized in natural language processing and beyond. In computer vision, Transformers have demonstrated outstanding capabilities across a wide range of computer vision tasks, such as image recognition \cite{dosovitskiy2020image,liu2021swin,touvron2021training}, object detection \cite{carion2020end,zhu2020deformable} and image segmentation \cite{ranftl2021vision,cheng2022masked,xie2021segformer}.
\citet{mondal2023gazeformer} introduced Gazeformer, a Transformer-based model specifically designed for zero-shot visual search scanpath prediction. In contrast, our proposed model is generic, capable of predicting both visual search and free-viewing scanpaths. Additionally, our model diverges from other Transformer-based architectures by drawing inspiration from the human vision system. It incorporates a novel foveation module simulating a simplified foveated retina, thereby establishing a dynamic visual working memory for enhanced scanpath prediction.

\section{Human Attention Transformer}
\label{sec:approach}
In this section, we first formulate scanpath prediction as a sequence of dense prediction tasks using behavior cloning. We then introduce our proposed transformer-based model, HAT, for scanpath prediction. Finally, we describe how we train HAT and use it for fast inference.

\subsection{Preliminaries}

To avoid the precision loss caused by grid discretization present in prior fixation prediction methods \cite{zelinsky2019benchmarking,yang2020predicting,chen2021predicting,yang2022target}, we formulate scanpath prediction as a sequential prediction of pixel coordinates. 
Given a $H{\times}W$ image and an optional initial fixation $f_0$ (often set as the center of an image), a scanpath prediction model  predicts a sequence of human-like fixation locations $f_1, \cdots, f_n$, with each fixation $f_i$ being a pixel location in the image. Note that $n$ is variable that may be different for each scanpath due to the different termination criteria of different human subjects. To model the uncertainty in human attention allocation, existing methods \cite{yang2022target,yang2020predicting,chen2021predicting,zelinsky2019benchmarking} often predict a probability distribution over a coarse grid of fixation locations at each step. HAT follows the same spirit but outputs a dense fixation heatmap. Specifically, HAT outputs a heatmap $Y_i\in[0,1]^{H{\times} W}$ with each pixel value indicating the chance of the pixel being fixated in the next fixation. In addition, HAT also outputs a termination probability $\tau_i\in[0,1]$ indicating how likely the model is to terminate the scanpath at the current step $i$. To sample a fixation, we apply $L_1$-normalization on $Y_i$. In the following, we omit the subscript $i$ for brevity.

\subsection{Network Architecture}
\begin{figure}[t]
  \centering
  \includegraphics[width=1.0\linewidth]{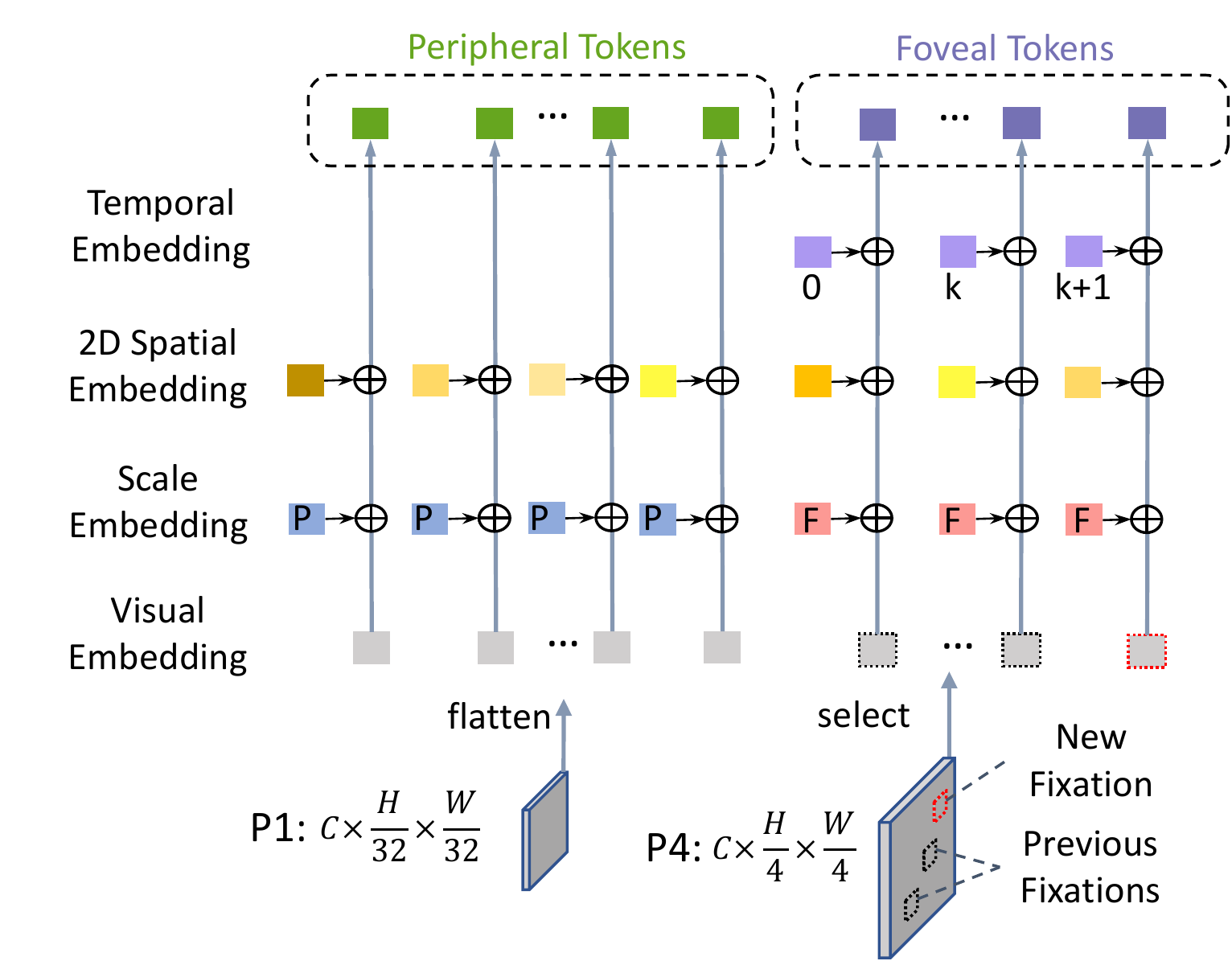}
  \caption{\textbf {Working memory construction.} We construct the working memory by starting with the visual embeddings (``what") flattened from $P_1$ over the spatial axes and selected from $P_4$ at previous fixation locations. A scale embedding is introduced to capture scale information. Spatial embeddings and temporal embeddings are further added to the tokens to enhance the ``where" and ``when" signals. At every new fixation (marked in red), we simply add a new foveal token while keeping other tokens unchanged.}
  \label{fig:foveation}
\end{figure}
HAT is a novel transformer-based model for scanpath prediction. At each fixation, HAT outputs a set of prediction pairs $\{(Y_t, \tau_t)\}_{t=1}^T$ where $t$ indicates a task, which could be a visual search task (e.g., clock search and mouse search) or a free-viewing task.
\Cref{fig:overview} shows an overview of the proposed model. HAT consists of four modules: 1) a feature extraction module that extracts a feature pyramid with multi-resolutional feature maps corresponding to information extracted at different eccentricities \cite{yang2022target,rashidi2020optimal}; 2) a foveation module which maintains a dynamical working memory representing the information acquired through fixations; 3) an aggregation module that selectively aggregates the information in the working memory using attention mechanism for each task; 4) a fixation prediction module that predicts the fixation heatmap $Y_t$ and termination probability $\tau_t$ for each task $t$.\\
\textbf{The feature extraction module}
consists of a pixel encoder (e.g., ResNet \cite{he2016deep}, a Swin transformer \cite{liu2021swin}), and a pixel decoder (e.g., FPN \cite{lin2017feature} and deformable attention \cite{zhu2020deformable}). Taking a $H{\times} W$ image as input, the pixel encoder encodes the input image into a high-semantic but low-resolution feature map. 
The pixel decoder up-samples the feature map several times, each time by a scale factor of two, to construct a pyramid of four multi-scale feature maps 
denoted as $P=\{P_1, \cdots, P_4\}$, where $P_1\in \mathbb{R}^{C{\times}\frac{H}{32}{\times}\frac{W}{32}}$, $P_4\in \mathbb{R}^{C{\times}\frac{H}{4}{\times}\frac{W}{4}}$, and $C$ is the channel dimension.\\
\textbf{The foveation module}
constructs a {\it dynamic} working memory using the feature maps $P_1$ and $P_4$ to represent the information a person acquires from the peripheral and foveal vision, respectively. We discard medium-grained feature maps $P_2$ and $P_3$ in computing the peripheral representation for computational efficiency. 
Finally, we apply a Transformer encoder \cite{vaswani2017attention} to dynamically update the working memory with the information acquired at a new fixation.
\Cref{fig:foveation} illustrates the construction of the working memory. 
The working memory consists of two parts: peripheral tokens and foveal tokens. We first flatten the low-resolution feature map $P_1$ over the spatial axes to obtain the peripheral visual embeddings $V^p\in\mathbb{R}^{(\frac{H}{32}\cdot\frac{W}{32}){\times} C}$. Feature vectors in $P_4$ at each fixation location are selected as the foveal visual embeddings $V^f\in\mathbb{R}^{k{\times} C}$, where $k$ is number of previous fixations. For simplicity, we round the fixation to its nearest position in $P_4$.
Then we add a learnable {\bf scale} embedding to each token to discern the scale/resolution of the visual embeddings. As the spatial information is shown to be important in predicting human scanpath (e.g., center bias and inhibition of return \cite{wang2010searching}), we enrich the peripheral and foveal tokens with their 2D {\bf spatial} information in the image. 
Specifically, we create a lookup table of 2D sinusoidal position embeddings \cite{li2021learnable} $G\in\mathbb{R}^{H{\times} W{\times} C}$ by concatenating the 1D sinusoidal positional encoding of the horizontal and vertical coordinates of each pixel location. For a visual embedding at position $(i, j)$ of a given feature map of stride $S$ ($S=32$ for $P_1$ and $S=4$ for $P_4$), its position encoding is defined by the element at position $(t_i, t_j)$ in $G$ where $t_i=\lfloor i\cdot S\rfloor$ and $t_j=\lfloor j\cdot S\rfloor$. 
Furthermore, we add to each foveal token the {\bf temporal} embedding, a learnable vector, according to its fixation index to capture the temporal order among previous fixations.\\
\textbf{The aggregation module}
is a transformer decoder \cite{vaswani2017attention} that selectively aggregates information from the working memory using a set of learnable, task-specific queries $Q\in \mathbb{R}^{N{\times} C}$, where $N$ is the number of tasks (e.g., $N=18$ for COCO-Search18 \cite{chen2021coco} and $N=1$ for free-viewing datasets). The transformer decoder has $L$ layers, with each layer consisting of a cross-attention layer, a self-attention layer and a feed-forward network (FFN). Different from the standard transformer decoder \cite{vaswani2017attention}, we follow \cite{cheng2022masked} and switch the order of cross-attention and self-attention module.
Firstly, each task query selectively gathers the information in working memory acquired through previous fixations using cross-attention. Then, the self-attention layer followed by a FFN is applied to exchange information in different queries which could boost the contextual cues \cite{chun1998contextual} in each query. When generating a scanpath, HAT maintains its state across fixations \textit{only} in the working memory, the input $Q$ is the same at each fixation prediction.\\
\textbf{The fixation prediction module} yields the final prediction---a fixation heatmap $\hat{Y}_t$ and a termination probability $\hat{\tau}_t$ for each task $t$. For the termination prediction, a linear layer followed by a sigmoid activation is applied on top of each updated query $q_t\in Q$:
\begin{equation}
    \hat{\tau}_t = \text{sigmoid}(Wq_t^T+b),
\end{equation}
where $W$ and $b$ are the parameters of the linear layer. For the fixation heatmap prediction, a Multi-Layer Perceptron (MLP) with two hidden layers first transforms $q_t$ into a task embedding, which is then convolved with the high-resolution feature map $P_4$ to get the fixation heatmap $\hat{Y}_t$ after a sigmoid layer:
\begin{equation}\label{eq:heatmap}
    \hat{Y}_t=\text{sigmoid}(P_4 \odot \text{MLP}(q_t)),
\end{equation}
where $\odot$ denotes the pixel-wise dot product operation. Finally, we upsample $\hat{Y}_t$ to the image resolution.
Note that the predictions for all tasks, i.e., $\hat{Y}\in\mathbb{R}^{N{\times} H{\times} W}$ and $\hat{\tau}\in\mathbb{R}^{N{\times}1}$, are yielded in parallel.

\subsection{Training and Inference}
\textbf{Training loss.}
We follow \cite{zelinsky2019benchmarking} and use behavior cloning to train HAT. The problem of scanpath prediction is broken down into learning a mapping from the input triplet of an image, a sequence of previous fixations, and a task to the output pair of a fixation heatmap and a termination probability. Given the predicted fixation heatmaps $\hat{Y}\in\mathbb{R}^{N{\times} H{\times} W}$ and termination probabilities $\hat{\tau}\in\mathbb{R}^{N{\times}1}$, the training loss is only calculated for its ground-truth task $t$:
\begin{equation}
    \mathcal{L}=\mathcal{L}_\text{fix}(\hat{Y}_t, Y)+\mathcal{L}_\text{term}(\hat{\tau}_t, \tau),
\end{equation} 
where $Y\in[0,1]^{H{\times} W}$ and $\tau\in\{0,1\}$ are the ground-truth fixation heatmap and termination label for task $t$, respectively. We compute $Y$ by smoothing the ground-truth fixation map with a  Gaussian kernel with the kernel size being one degree of visual angle. $\mathcal{L}_\text{fix}$ denotes the fixation loss and is computed using pixel-wise focal loss \cite{lin2017focal,law2018cornernet}:
\begin{equation}
\begin{aligned}
\mathcal{L}_\text{fix} = \frac{-1}{HW}\sum_{i,j}
\begin{cases} 
    (1-\hat{Y}_{ij})^\alpha\log(\hat{Y}_{ij}) & \textrm{if } Y_{ij}=1,\\[5pt]
    \begin{gathered}
    (1-{Y}_{ij})^\beta(\hat{Y}_{ij})^\alpha\\
    \log(1-\hat{Y}_{ij}) 
    \end{gathered}& \text{otherwise},
\end{cases}
\end{aligned}
\label{eq:loss_fn}
\end{equation}
where $Y_{ij}$ represents the value of $Y$ at location $(i, j)$ and we set $\alpha=2$ and $\beta=4$ following \cite{yang2022target,law2018cornernet}.
$\mathcal{L}_\text{term}$ is the termination loss and is computed by applying a binary cross entropy (negative log-likelihood) loss, i.e., 
\begin{equation}
    \mathcal{L}_\text{term}=-\omega\cdot\tau\log(\hat{\tau}_t)- (1-\tau)\log(1-\hat{\tau}_t),
\end{equation}
where $\omega$ is a weight to balance the loss of positive and negative training examples since there are many more  negative labels than positive labels for training a termination prediction, especially for target-absent visual search and free-viewing tasks where scanpath are long. We set $\omega$ to be the ratio of the number of negative training instances to the number of positive ones.\\
\textbf{Inference.}
Similarly to \cite{yang2020predicting,yang2022target,chen2021predicting}, HAT also generates scanpaths autoregressively, but in an efficient way. Given an image, HAT only computes the image pyramid $P$ and peripheral tokens once. For a new fixation, a foveal token is constructed and appended to the working memory after which the aggregation module and fixation prediction module yield the fixation heatmaps and termination predictions for all tasks in parallel.

\section{Experiments}
\label{sec:experiments}
\textbf{Datasets.}
We train and evaluate HAT using four datasets: COCO-Search18 \cite{chen2021coco}, COCO-FreeView \cite{Chen_2022_CVPR}, MIT1003 \cite{judd2009learning} and OSIE \cite{xu2014predicting}. COCO-Search18 is a large-scale visual search dataset containing human scanpaths in searching for 18 different object target and it has two parts: target-present and target-absent. In total, there are 3101 target-present images and 3101 target-absent images in COCO-Search18, each viewed by 10 subjects. Following \cite{yang2022target}, we treat the target-present part and target-absent part of COCO-Search18 as two separate datasets and train models on them independently.
COCO-FreeView is a ``sibling" dataset of COCO-Search18 but with free-viewing scanpaths. COCO-FreeView contains the same images with COCO-Search18, each viewed by 10 subjects in a free-viewing setting. 
MIT1003 is a widely-used free-viewing dataset containing 1003 natural images. OSIE is a free-viewing gaze dataset with rich semantic-level annotations, containing 700 natural indoor and outdoor images. Each image in MIT1003 and OSIE is viewed by 15 subjects.\\
\textbf{Evaluation metrics.}
To measure the performance, we mainly analyze the scanpath prediction models from two aspects: 1) how similar the predicted scanpaths are to the  human scanpaths; and 2) how accurate a model predicts the next fixation {\it given all previous fixations}. 
To measure the scanpath similarity, we use a commonly adopted metric, {sequence score (SS)} \cite{borji2013analysis} and its variant {semantic sequence score (SemSS)} \cite{yang2022target}. SS transforms the scanpaths into sequences of fixation cluster IDs and then compares them using a string matching algorithm \cite{needleman1970general}. Different from SS, SemSS transforms a scanpath into a string of semantic labels of the fixated pixels. 
For next fixation prediction, we follow \cite{kummerer2021state,yang2022target,kummerer2022deepgaze} and report the conditional saliency metrics, {cIG, cNSS and cAUC}, which measure how well a predicted fixation probability map of a model predicts the ground-truth (next) fixation when the model is provided with the fixation history of the scanpath in consideration, using the widely used saliency metrics, IG, NSS and AUC \cite{bylinskii2018different}. For fair comparison, we follow \cite{yang2022target} and predict one scanpath for each testing image, step by step selecting the most probable fixation location as the next fixation.\\
\textbf{Baselines.}
We first compare our model against several heuristic baselines. Following prior works \cite{yang2020predicting,zelinsky2019benchmarking,yang2022target,chen2021predicting,kummerer2022deepgaze}, the {human consistency}, an oracle where we use one viewer's scanpath to predict the scanpath of another, is reported as a gold-standard model. Second, we compare to a {fixation heuristic} method---a ConvNet trained to predict human fixation density maps, from which we select fixations sequentially with inhibition of return. For visual search scanpaths, we further include a {detector} baseline, which is similar to the fixation heuristic, but trained on target-present images of COCO-Search18 to predict a target detection probability map. For both fixation heuristic and detector baselines, we use the winner-take-all strategy to generate scanpaths.
Furthermore, we compare HAT to the previous state-of-the-art models of scanpath prediction: {IVSN} \cite{zhang2018finding}, PathGAN \cite{assens2018pathgan}, {IRL} \cite{yang2020predicting}, {Chen \etal} \cite{chen2021predicting}, DeepGaze III \cite{kummerer2022deepgaze}, {FFMs} \cite{yang2022target} and GazeFormer \cite{mondal2023gazeformer}.
Note that IVSN only applies for visual search tasks, and unlike other methods, IVSN is designed for zero-shot search scanpath prediction, hence is not trained with any gaze data. DeepGaze III only applies for free-viewing scanpaths and is trained with the SALICON dataset \cite{jiang2015salicon} and MIT1003 \cite{judd2009learning}.\\
\textbf{Implementation details.}
We use ResNet-50 \cite{he2016deep} as the pixel encoder and MSDeformAttn \cite{zhu2020deformable} as the pixel decoder. For the foveation module, the transformer encoder has three layers. The transformer decoder in the aggregation module has six layers (i.e,. $L=6$). All transformer encoder and decoder layers in HAT have 4 attention heads. The MLP in the fixation prediction module has two linear layers with 512 hidden dimensions and a ReLU activation function.
We use the AdamW \cite{loshchilov2017decoupled} with the learning rate of $0.0001$ and train HAT for 30 epochs with a batch size of 128. All images are resized to $320{\times} 512$ for computational efficiency during training and inference. Following \cite{yang2020predicting}, we set the maximum length of each predicted scanpath to 6 and 10 (excluding the initial fixation) for target-present and target-absent search scanpath prediction, respectively. For free viewing, the maximum scanpath length is set to 20.
For more implementation details, please refer to the supplement.

\setlength{\tabcolsep}{2.5pt}
\begin{table}[t]
\begin{center}
\begin{tabular}{l|ccccc}
\toprule
 &  SemSS  & SS & cIG  & cNSS & cAUC\\ 
\midrule
Human consistency & 0.500 & 0.500 & - & - & -\\\midrule
Detector & 0.523 & 0.449 & 0.182 & 2.346 & 0.905\\
Fixation heuristic & 0.506 & 0.437&1.107&	2.186 &  0.917\\
IVSN \cite{zhang2018finding} & 0.368 & 0.326 & -0.192 & 1.318 & 0.901\\
PathGAN \cite{assens2018pathgan} & 0.280 & 0.239 & - & - & -\\
IRL \cite{yang2020predicting} & 0.486 & 0.422 & -9.709 & 1.977 & 0.913\\
Chen~\etal \cite{chen2021predicting} & 0.518 & 0.445 & -1.273 &2.606 & 0.956\\
FFMs \cite{yang2022target} & 0.500 & {0.451} & 1.548 & 2.376 & 0.932 \\
Gazeformer \cite{mondal2023gazeformer} & 0.499 & \bf{0.489} & - & - & -\\
\textbf{HAT} (ours) & \bf{0.543} & 0.470 & \bf{2.399} & \bf{5.086} & \bf{0.977}\\
\bottomrule
\end{tabular}
\caption{{\bf Target-present search scanpath prediction comparison} on the {target-present} test set of COCO-Search18. We highlight the best results in bold.}
\label{tb:tp_rst}
\end{center}
\vspace{-0.35cm}
\end{table}
\setlength{\tabcolsep}{2.5pt}
\begin{table}[t]
\begin{center}
\begin{tabular}{l|ccccc}
\toprule
 &  SemSS  & SS & cIG  & cNSS & cAUC\\ 
\midrule
Human consistency & 0.372 & 0.381 & - & - & -\\\midrule
Detector & 0.332 & 0.321 & -0.516 & 0.446 & 0.783\\
Fixation heuristic & 0.309 & 0.298	&	-0.599&	0.405 & 0.798\\
IVSN \cite{zhang2018finding} & 0.279 & 0.260 & -0.219 & 0.884 & 0.867\\
PathGAN \cite{assens2018pathgan} & 0.315 & 0.250 & - & - & -\\
IRL \cite{yang2020predicting} & 0.329 & 0.319 & 0.032 & 1.202& 0.893\\
Chen~\etal \cite{chen2021predicting} & 0.340 & 0.331 & -3.278 & 1.600& 0.925\\
FFMs \cite{yang2022target} & 0.376 & 0.372 & 0.729 & 1.524 & 0.916 \\
Gazeformer \cite{mondal2023gazeformer} & 0.374 & 0.357 & - & - & -\\
\textbf{HAT} (ours) & \bf{0.382} & \bf{0.402} & \bf{1.686} & \bf{3.103} & \bf{0.961}\\
\bottomrule
\end{tabular}
\caption{{\bf Target-absent search scanpath prediction comparison} on the {target-absent} test set of COCO-Search18. We highlight the best results in bold.}
\label{tb:ta_rst}
\end{center}
\vspace{-0.35cm}
\end{table}
\setlength{\tabcolsep}{5pt}
\begin{table}[t]
\begin{center}
\begin{tabular}{l|cccccccc}
\toprule
& SS & cIG  & cNSS & cAUC\\
\midrule
Human consistency & 0.349  & - & - & -\\\midrule
Fixation heuristic & 0.329 & 0.319 & 1.621 & 0.930\\
PathGAN \cite{assens2018pathgan} & 0.181 & - & - & -\\
IRL \cite{yang2020predicting} & 0.300 & -0.213 & 1.018 & 0.888\\
Chen \etal \cite{chen2021predicting} & {0.365} & -1.263 & 1.655 & 0.922\\
DeepGaze III \cite{kummerer2022deepgaze} & 0.339 & 0.140 & 1.418 & 0.910\\
FFMs \cite{yang2022target} &	0.329&	0.329&	1.432 & 0.918\\
Gazeformer \cite{mondal2023gazeformer} & 0.280 & - & - & - \\
HAT & {\bf 0.369} & {\bf 1.485} & {\bf 3.382} & {\bf 0.965}\\
\bottomrule
\end{tabular}
\caption{{\bf Comparing free-viewing scanpath prediction algorithms} (rows) using multiple metrics (columns) on the test set of COCO-FreeView. The best results are highlighted in bold.}
\label{tb:fv_rst}
\end{center}
\vspace{-0.35cm}
\end{table}

\subsection{Main Results}
\textbf{Target-present search.}
We compare HAT with previous scanpath prediction models under the target-present (TP) setting using the target-present part of the COCO-Search18 dataset in \cref{tb:tp_rst}. HAT consistently outperforms all other predictive methods in predicting TP human scanpaths in nearly all metrics. The simple heuristic baselines (i.e., detector and fixation heuristic) perform quite well on TP scanpath prediction by predicting the location of the target or fixation density map as in 60\% of the TP trials of COCO-Search18 humans can locate the target within 2 fixations. However, they have low scores on saliency metrics (i.e., cIG, cNSS and cAUC) as they ignore the inter-dependencies between fixations. Compared to FFMs \cite{yang2022target} and Chen \etal \cite{chen2021predicting} which have high saliency scores, HAT further improves the performance significantly for all metrics. Particularly, HAT is better than Chen \etal \cite{chen2021predicting} (the second best) in cNSS by 95\%. HAT slightly lags behind the most recent GazeFormer \cite{mondal2023gazeformer} in SS but is significantly better in semSS. We also demonstrate in the supplement that HAT learns the entire scanpath distribution from multiple subjects whereas GazeFormer overfits to the \enquote{average person} and fails to predict scanpaths from different subjects. 
Moreover, HAT surpasses the human consistency in semSS, suggesting that HAT well captures the semantics behind fixations.\\
\textbf{Target-absent search.}
For target-absent (TA) search scanpath prediction, we compare HAT to different approaches on the TA test set of COCO-Search18 in \cref{tb:ta_rst}. Different from TP search results shown in \cref{tb:tp_rst}, we see in \cref{tb:ta_rst} that the gap between heuristic methods to human consistency is much larger for TA search, demonstrating that TA search scanpath prediction is a more challenging task than TP scanpath prediction. Indeed, the predominant influence on human attention in TP search (i.e., the target) is now absent \cite{Chen_2022_CVPR}, making other factors such as the spatial cues provided by the anchor objects \cite{boettcher2018anchoring}, the contextual cues from global scene understanding \cite{torralba2006contextual} and object co-occurrence \cite{mack2011object} stand out. The discernment of these factors necessitates a robust semantic understanding of the input image. \cref{tb:ta_rst} shows that HAT sets a new state-of-the-art at \textbf{all} metrics, outperforming the previous state-of-the-art (Chen \etal \cite{chen2021predicting}) by 94\% in cNSS. More importantly, HAT achieves a sequence score surpassing human consistency for the first time. These results suggest that comparing to other methods HAT better captures the semantics of the image and learns the relation between other objects and targets.
\\
\textbf{Free-viewing.}
In addition to visual search, HAT can predict free-viewing scanpaths by treating free-viewing as a standalone task. In \cref{tb:fv_rst}, we compare HAT with the baselines using COCO-FreeView. Note that Detector and IVSN are excluded here as the free-viewing fixations are not tasked to searching for a target like visual search. HAT outperforms all other methods in cIG, cNSS and cAUC, especially HAT is 351\% and 104\% better than the second best (FFMs and Chen \etal \cite{chen2021predicting}) in cIG, cNSS, respectively. This reaffirms the effectiveness of HAT as a generic framework for scanpath prediction. We further validated the effectiveness of our proposed HAT using OSIE \cite{xu2014predicting} and MIT1003 \cite{judd2009learning}, and the  generalizability of HAT to new scenes, please refer to the supplement for detailed results.




\subsection{Qualitative Analysis}\label{sec:qual}

\begin{figure}[t]
  \centering
  \includegraphics[width=1.0\linewidth]{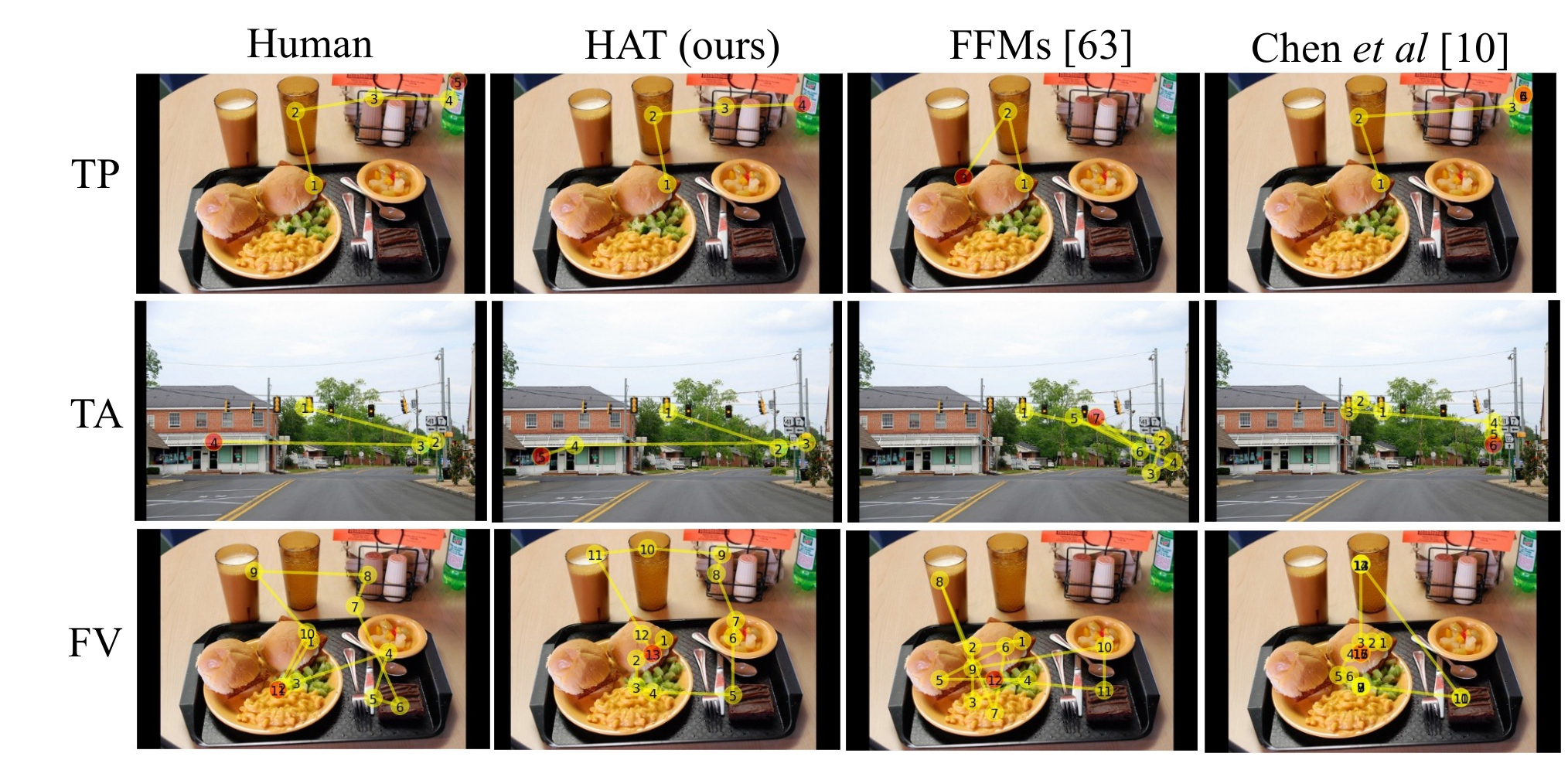}
  \caption{\textbf{Visualization of the ground-truth human scanpaths
and predicted scanpaths of different methods (columns).} Three different settings (rows) including target-present bottle search, target-absent stop sign search and free viewing are shown from the top to bottom. The final fixation of each scanpath is highlighted in red circle. For methods without termination prediction, i.e., IRL, detector and fixation heuristic, we visualize the first 6 fixations for visual search and 15 for free viewing. The rightmost column shows the predicted scanpaths of the heuristic methods (detector 630 for visual search and fixation heuristic for free-viewing)}
  \label{fig:vis_small}
\end{figure}
\textbf{Scanpath visualization.}
In this section, we qualitatively compare the predicted scanpaths of different methods to each other and to the ground-truth human scanpaths in the TP, TA and FV settings. As shown in \cref{fig:vis_small}, when searching for bottles in the TP setting, HAT not only correctly predicted the terminal fixation on the heavily-occluded target, but also predicted fixations on all the distractor objects that look similar to the target, like humans do. Other methods either missed the distractor objects or failed to find the target. Similarly, for the TA stop sign search, HAT was the only one that looked at both sides of the road in searching for a stop sign like the human subject would, showing a use of semantic and context cues to control attention. In the FV setting, HAT also predicted the most human-alike scanpaths among all methods in (1) the fixation locations (where), (2) the semantics (what), and (3) the order (when) of the fixations. More scanpath visualizations can be found in the supplement.\\
\begin{figure}[t]
  \centering
{\includegraphics[width=1.0\linewidth]{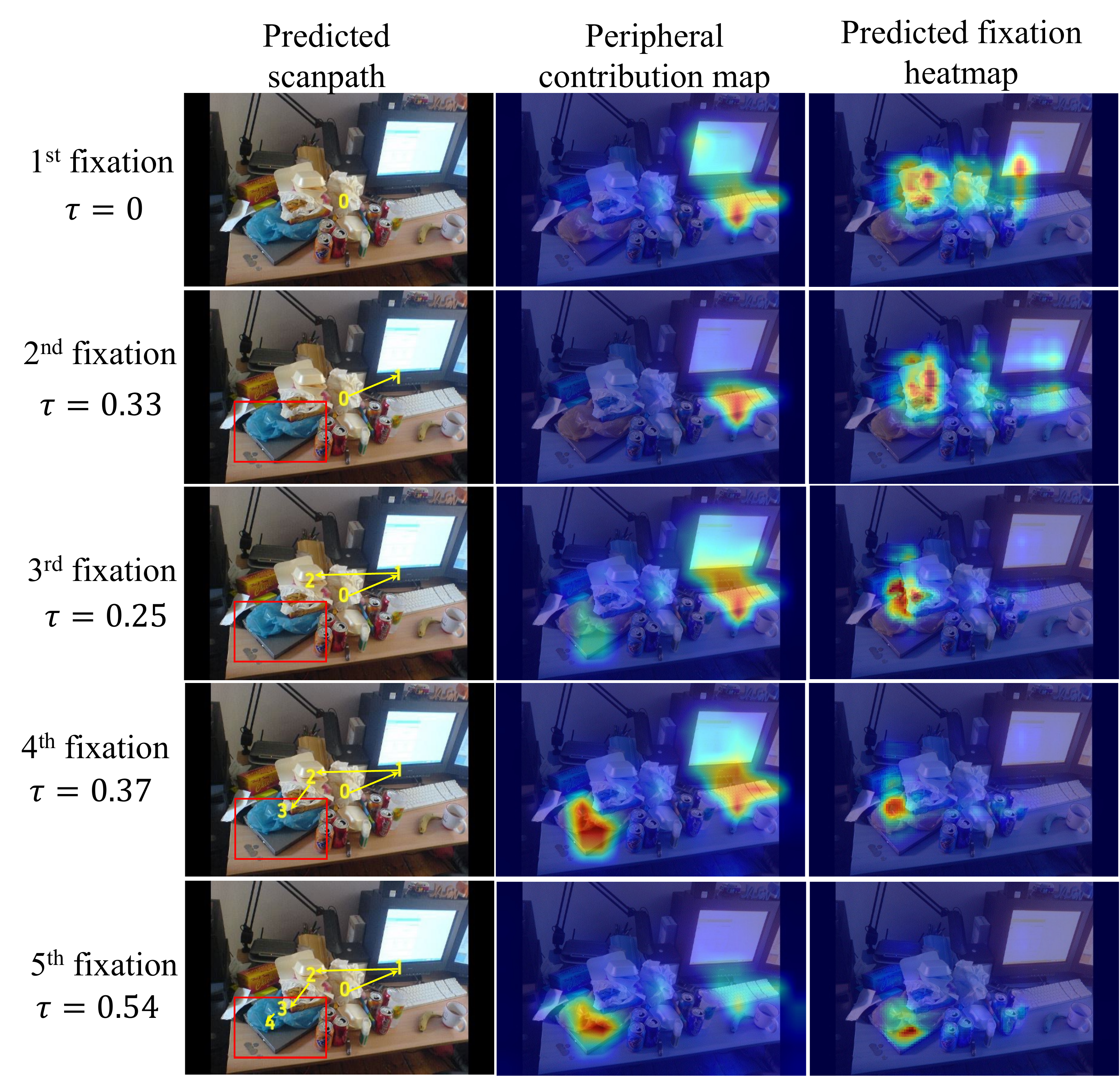}\label{fig:tp_peri_cont}}
\caption{Visualization of the {\bf predicted scanpath, peripheral contribution map and fixation heatmap} (columns) of HAT for target-present laptop visual search examples at every fixation (rows). We also include the predicted termination probability $\tau$ for each step on the left. The model terminates searching if $\tau>0.5$.}
\label{fig:peri_cont}
\vspace{-0.35cm}
\end{figure}
\\
\textbf{Model interpretability.}
A distinctive attribute of HAT lies in its interpretability, facilitated by the computational attention mechanism and the foveation module design. HAT enables quantitatively measuring the contribution of both peripheral and foveal tokens to fixation allocation. The contribution of a token is computed as the attention
weight from the last cross-attention layer of the aggregation module in HAT. By computing the normalized contribution of each peripheral token, we create a \textit{peripheral contribution map}, which offers insights for the human gaze behavior.
We further analyze how the peripheral contribution map evolves across a sequence of fixations. \cref{fig:peri_cont} shows the predicted scanpath, peripheral contribution maps and predicted fixation heatmaps of HAT in a TP laptop search task. We observe that the encoded periphery features not only align with the location of the next fixation (e.g., when the occluded laptop is encoded in the left-bottom periphery, the model makes a fixation to the target and terminates the search), but also provides the {\it contextual cues} where a target might be located (e.g., near the keyboard and the monitor where a laptop is usually found). We also observe a similar pattern for the TA setting (see illustration in the supplement). In the supplement, we also collectively analyze the contribution of peripheral and foveal tokens in predicting human attention control, which has shown that the peripheral vision plays different roles under different settings. These all have demonstrated that HAT can make highly \textit{interpretable} predictions.\\

\subsection{Ablation studies}\label{sec:ablation}
We ablate HAT under the TA setting as TA search fixations exhibit characteristics of both target-present search fixations and free-viewing fixations \cite{Chen_2022_CVPR}.\\
\textbf{Peripheral and foveal tokens.}
We verify the effectiveness of peripheral tokens and foveal tokens by ablating them one at a time. It is shown in \cref{tb:ablation} that ablating any one of them incur a performance drop over all metrics. This suggests that all of these components contribute to the superior performance of HAT. In comparison, removing foveal tokens incurs a larger performance drop (cIG decreases by 30\%). This decline is expected as foveal tokens embody the knowledge accumulated from prior fixations. Without them, HAT can be regarded as a static fixation density map predictor, akin to the fixation heuristic baseline. Conversely, the removal of  peripheral tokens has a relatively minor effect, possibly attributed to the adaptive capacity of foveal tokens ($P_2$) compensating for information loss in peripheral tokens ($P_1$) during training.\\
\textbf{Output resolution.}
HAT has a default output resolution of $80\times128$ due to the convolution with the high-resolution feature map $P_4$ (see \cref{fig:overview}). In \cref{tb:ablation} (last row), we change the convolution operant from $P_4$ to $P_2$ to yield an output resolution of $20\times 32$, same as FFMs \cite{yang2022target} and IRL \cite{yang2020predicting} but smaller than Chen~\etal \cite{chen2021predicting} ($30\times40$). 
Despite that a reduced resolution incurs a noticeable performance drop in HAT, HAT still outperforms prior state-of-the-art FFMs with the same output resolution and \citet{chen2021predicting} using a higher output resolution. This underscores HAT's effectiveness and design flexibility. Additional ablations can be found in the supplement.

\setlength{\tabcolsep}{2.5pt}
\begin{table}[t]
\begin{center}
\begin{tabular}{l|ccccc}
\toprule 
 &  SemSS  & SS & cIG & cNSS & cAUC\\ 
\midrule
baseline ($\htimesw{80}{128}$) & \bf{0.382} & \bf{0.402} & \bf{1.686} & \bf{3.103} & \bf{0.961}\\
\hline
-- peripheral tokens & 0.375 & 0.396 & 1.600 & 3.003 & 0.960\\
-- foveal tokens & 0.358 & 0.385 & 1.179 & 2.380 & 0.948\\
\hline
low-res ($\htimesw{20}{32}$) & 0.374 & 0.389 & 1.534 & 2.760 & 0.955\\
\bottomrule
\end{tabular}
\caption{{\bf Ablation study} of HAT. These experiments are done on the TA set of COCO-Search18. The best results are in bold.}
\label{tb:ablation}
\end{center}
\vspace{-0.35cm}
\end{table}

\section{Conclusions}
\label{sec:conclusions}
With the rapid development of Augmented Reality (AR) and Virtual Reality (VR) technologies, there is an increasing demand for predicting and understanding human gaze behavior \cite{park2021mosaic,kaplanyan2019deepfovea,park2019advancing}, with scanpath prediction being a challenging task. For those AR/VR applications requiring a high input resolution ($360^\circ$), discretizing fixations into a coarse grid incurs a non-negligible loss in accuracy.
In this work we presented HAT, a generic attention scanpath prediction model. Built from a simple dense prediction framework \cite{cheng2021per}, HAT circumvents the drawbacks of discretizing fixations as in prior state of the arts \cite{yang2020predicting,chen2021predicting,yang2022target}. Inspired by the human vision system, HAT uses a novel foveated working memory which dynamically updates its knowledge about the scene as it changes its fixation. We show that HAT achieves new SOTA performance, not only in predicting free-viewing fixation scanpaths, but also scanpaths in target-present and target-absent search. In demonstrating this broad scope, our HAT model sets a new bar in the computational attention of attention control. 


\myheading{Acknowledgement}. {The authors would like to thank Xianyu Chen for providing the source code for PathGAN. This project was supported by US National Science Foundation Awards IIS-1763981, IIS-2123920, NSDF DUE-2055406, and the SUNY2020 Infrastructure Transportation Security Center, and a gift from Adobe.}

{
    \small
    \bibliographystyle{ieeenat_fullname}
    \bibliography{main}

\begin{thebibliography}{69}
\providecommand{\natexlab}[1]{#1}
\providecommand{\url}[1]{\texttt{#1}}
\expandafter\ifx\csname urlstyle\endcsname\relax
  \providecommand{\doi}[1]{doi: #1}\else
  \providecommand{\doi}{doi: \begingroup \urlstyle{rm}\Url}\fi

\bibitem[Assens et~al.(2018)Assens, Giro-i Nieto, McGuinness, and O'Connor]{assens2018pathgan}
Marc Assens, Xavier Giro-i Nieto, Kevin McGuinness, and Noel~E O'Connor.
\newblock Pathgan: Visual scanpath prediction with generative adversarial networks.
\newblock In \emph{ECCV Workshops}, 2018.

\bibitem[Assens~Reina et~al.(2017)Assens~Reina, Giro-i Nieto, McGuinness, and O'Connor]{assens2017saltinet}
Marc Assens~Reina, Xavier Giro-i Nieto, Kevin McGuinness, and Noel~E O'Connor.
\newblock Saltinet: Scan-path prediction on 360 degree images using saliency volumes.
\newblock In \emph{ICCV Workshops}, 2017.

\bibitem[Berg et~al.(2009)Berg, Boehnke, Marino, Munoz, and Itti]{berg2009free}
David~J Berg, Susan~E Boehnke, Robert~A Marino, Douglas~P Munoz, and Laurent Itti.
\newblock Free viewing of dynamic stimuli by humans and monkeys.
\newblock \emph{Journal of vision}, 9\penalty0 (5):\penalty0 19--19, 2009.

\bibitem[Boettcher et~al.(2018)Boettcher, Draschkow, Dienhart, and V{\~o}]{boettcher2018anchoring}
Sage~EP Boettcher, Dejan Draschkow, Eric Dienhart, and Melissa L-H V{\~o}.
\newblock Anchoring visual search in scenes: Assessing the role of anchor objects on eye movements during visual search.
\newblock \emph{Journal of vision}, 18\penalty0 (13):\penalty0 11--11, 2018.

\bibitem[Borji and Itti(2013)]{borji2013state}
Ali Borji and Laurent Itti.
\newblock State-of-the-art in visual attention modeling.
\newblock \emph{IEEE transactions on pattern analysis and machine intelligence}, 35\penalty0 (1):\penalty0 185--207, 2013.

\bibitem[Borji et~al.(2013)Borji, Tavakoli, Sihite, and Itti]{borji2013analysis}
Ali Borji, Hamed~R Tavakoli, Dicky~N Sihite, and Laurent Itti.
\newblock Analysis of scores, datasets, and models in visual saliency prediction.
\newblock In \emph{ICCV}, 2013.

\bibitem[Borji et~al.(2015)Borji, Cheng, Jiang, and Li]{borji2015salient}
Ali Borji, Ming-Ming Cheng, Huaizu Jiang, and Jia Li.
\newblock Salient object detection: A benchmark.
\newblock \emph{IEEE transactions on image processing}, 24\penalty0 (12):\penalty0 5706--5722, 2015.

\bibitem[Bylinskii et~al.(2018)Bylinskii, Judd, Oliva, Torralba, and Durand]{bylinskii2018different}
Zoya Bylinskii, Tilke Judd, Aude Oliva, Antonio Torralba, and Fr{\'e}do Durand.
\newblock What do different evaluation metrics tell us about saliency models?
\newblock \emph{IEEE transactions on pattern analysis and machine intelligence}, 41\penalty0 (3):\penalty0 740--757, 2018.

\bibitem[Carion et~al.(2020)Carion, Massa, Synnaeve, Usunier, Kirillov, and Zagoruyko]{carion2020end}
Nicolas Carion, Francisco Massa, Gabriel Synnaeve, Nicolas Usunier, Alexander Kirillov, and Sergey Zagoruyko.
\newblock End-to-end object detection with transformers.
\newblock In \emph{ECCV}, 2020.

\bibitem[Chen et~al.(2021{\natexlab{a}})Chen, Jiang, and Zhao]{chen2021predicting}
Xianyu Chen, Ming Jiang, and Qi Zhao.
\newblock Predicting human scanpaths in visual question answering.
\newblock In \emph{CVPR}, 2021{\natexlab{a}}.

\bibitem[Chen et~al.(2021{\natexlab{b}})Chen, Yang, Ahn, Samaras, Hoai, and Zelinsky]{chen2021coco}
Yupei Chen, Zhibo Yang, Seoyoung Ahn, Dimitris Samaras, Minh Hoai, and Gregory Zelinsky.
\newblock Coco-search18 fixation dataset for predicting goal-directed attention control.
\newblock \emph{Scientific reports}, 11\penalty0 (1):\penalty0 1--11, 2021{\natexlab{b}}.

\bibitem[Chen et~al.(2022)Chen, Yang, Chakraborty, Mondal, Ahn, Samaras, Hoai, and Zelinsky]{Chen_2022_CVPR}
Yupei Chen, Zhibo Yang, Souradeep Chakraborty, Sounak Mondal, Seoyoung Ahn, Dimitris Samaras, Minh Hoai, and Gregory Zelinsky.
\newblock Characterizing target-absent human attention.
\newblock In \emph{CVPR Workshops}, 2022.

\bibitem[Cheng et~al.(2021)Cheng, Schwing, and Kirillov]{cheng2021per}
Bowen Cheng, Alex Schwing, and Alexander Kirillov.
\newblock Per-pixel classification is not all you need for semantic segmentation.
\newblock In \emph{NeurIPS}, 2021.

\bibitem[Cheng et~al.(2022)Cheng, Misra, Schwing, Kirillov, and Girdhar]{cheng2022masked}
Bowen Cheng, Ishan Misra, Alexander~G Schwing, Alexander Kirillov, and Rohit Girdhar.
\newblock Masked-attention mask transformer for universal image segmentation.
\newblock In \emph{CVPR}, 2022.

\bibitem[Chun and Jiang(1998)]{chun1998contextual}
Marvin~M Chun and Yuhong Jiang.
\newblock Contextual cueing: Implicit learning and memory of visual context guides spatial attention.
\newblock \emph{Cognitive psychology}, 36\penalty0 (1):\penalty0 28--71, 1998.

\bibitem[Cornia et~al.(2018)Cornia, Baraldi, Serra, and Cucchiara]{cornia2018predicting}
Marcella Cornia, Lorenzo Baraldi, Giuseppe Serra, and Rita Cucchiara.
\newblock Predicting human eye fixations via an lstm-based saliency attentive model.
\newblock \emph{IEEE Transactions on Image Processing}, 27\penalty0 (10):\penalty0 5142--5154, 2018.

\bibitem[Desimone et~al.(1995)Desimone, Duncan, et~al.]{desimone1995neural}
Robert Desimone, John Duncan, et~al.
\newblock Neural mechanisms of selective visual attention.
\newblock \emph{Annual review of neuroscience}, 18\penalty0 (1):\penalty0 193--222, 1995.

\bibitem[Dosovitskiy et~al.(2020)Dosovitskiy, Beyer, Kolesnikov, Weissenborn, Zhai, Unterthiner, Dehghani, Minderer, Heigold, Gelly, et~al.]{dosovitskiy2020image}
Alexey Dosovitskiy, Lucas Beyer, Alexander Kolesnikov, Dirk Weissenborn, Xiaohua Zhai, Thomas Unterthiner, Mostafa Dehghani, Matthias Minderer, Georg Heigold, Sylvain Gelly, et~al.
\newblock An image is worth 16x16 words: Transformers for image recognition at scale.
\newblock In \emph{ICLR}, 2020.

\bibitem[Findlay and Gilchrist(2001)]{findlay2001visual}
John~M Findlay and Iain~D Gilchrist.
\newblock Visual attention: The active vision perspective.
\newblock In \emph{Vision and attention}, pages 83--103. Springer, 2001.

\bibitem[Gazzaley and Nobre(2012)]{gazzaley2012top}
Adam Gazzaley and Anna~C Nobre.
\newblock Top-down modulation: bridging selective attention and working memory.
\newblock \emph{Trends in cognitive sciences}, 16\penalty0 (2):\penalty0 129--135, 2012.

\bibitem[He et~al.(2016)He, Zhang, Ren, and Sun]{he2016deep}
Kaiming He, Xiangyu Zhang, Shaoqing Ren, and Jian Sun.
\newblock Deep residual learning for image recognition.
\newblock In \emph{CVPR}, 2016.

\bibitem[Huang et~al.(2015)Huang, Shen, Boix, and Zhao]{huang2015salicon}
Xun Huang, Chengyao Shen, Xavier Boix, and Qi Zhao.
\newblock Salicon: Reducing the semantic gap in saliency prediction by adapting deep neural networks.
\newblock In \emph{ICCV}, 2015.

\bibitem[Itti and Koch(2000)]{itti2000saliency}
Laurent Itti and Christof Koch.
\newblock A saliency-based search mechanism for overt and covert shifts of visual attention.
\newblock \emph{Vision research}, 40\penalty0 (10-12):\penalty0 1489--1506, 2000.

\bibitem[Jetley et~al.(2016)Jetley, Murray, and Vig]{jetley2016end}
Saumya Jetley, Naila Murray, and Eleonora Vig.
\newblock End-to-end saliency mapping via probability distribution prediction.
\newblock In \emph{CVPR}, 2016.

\bibitem[Jiang et~al.(2015)Jiang, Huang, Duan, and Zhao]{jiang2015salicon}
Ming Jiang, Shengsheng Huang, Juanyong Duan, and Qi Zhao.
\newblock Salicon: Saliency in context.
\newblock In \emph{CVPR}, 2015.

\bibitem[Judd et~al.(2009)Judd, Ehinger, Durand, and Torralba]{judd2009learning}
Tilke Judd, Krista Ehinger, Fr{\'e}do Durand, and Antonio Torralba.
\newblock Learning to predict where humans look.
\newblock In \emph{ICCV}, 2009.

\bibitem[Kaplanyan et~al.(2019)Kaplanyan, Sochenov, Leimk{\"u}hler, Okunev, Goodall, and Rufo]{kaplanyan2019deepfovea}
Anton~S Kaplanyan, Anton Sochenov, Thomas Leimk{\"u}hler, Mikhail Okunev, Todd Goodall, and Gizem Rufo.
\newblock Deepfovea: Neural reconstruction for foveated rendering and video compression using learned statistics of natural videos.
\newblock \emph{ACM Transactions on Graphics (TOG)}, 38\penalty0 (6):\penalty0 1--13, 2019.

\bibitem[Kruthiventi et~al.(2017)Kruthiventi, Ayush, and Babu]{kruthiventi2017deepfix}
Srinivas~SS Kruthiventi, Kumar Ayush, and R~Venkatesh Babu.
\newblock Deepfix: A fully convolutional neural network for predicting human eye fixations.
\newblock \emph{IEEE Transactions on Image Processing}, 26\penalty0 (9):\penalty0 4446--4456, 2017.

\bibitem[K{\"u}mmerer and Bethge(2021)]{kummerer2021state}
Matthias K{\"u}mmerer and Matthias Bethge.
\newblock State-of-the-art in human scanpath prediction.
\newblock \emph{arXiv preprint arXiv:2102.12239}, 2021.

\bibitem[K{\"u}mmerer et~al.(2014)K{\"u}mmerer, Theis, and Bethge]{kummerer2014deep}
Matthias K{\"u}mmerer, Lucas Theis, and Matthias Bethge.
\newblock Deep gaze i: Boosting saliency prediction with feature maps trained on imagenet.
\newblock \emph{arXiv preprint arXiv:1411.1045}, 2014.

\bibitem[Kummerer et~al.(2017)Kummerer, Wallis, Gatys, and Bethge]{kummerer2017understanding}
Matthias Kummerer, Thomas~SA Wallis, Leon~A Gatys, and Matthias Bethge.
\newblock Understanding low-and high-level contributions to fixation prediction.
\newblock In \emph{ICCV}, 2017.

\bibitem[K{\"u}mmerer et~al.(2022)K{\"u}mmerer, Bethge, and Wallis]{kummerer2022deepgaze}
Matthias K{\"u}mmerer, Matthias Bethge, and Thomas~SA Wallis.
\newblock Deepgaze iii: Modeling free-viewing human scanpaths with deep learning.
\newblock \emph{Journal of Vision}, 22\penalty0 (5):\penalty0 7--7, 2022.

\bibitem[Land and Tatler(2009)]{land2009looking}
Michael Land and Benjamin Tatler.
\newblock \emph{Looking and acting: vision and eye movements in natural behaviour}.
\newblock Oxford University Press, 2009.

\bibitem[Law and Deng(2018)]{law2018cornernet}
Hei Law and Jia Deng.
\newblock Cornernet: Detecting objects as paired keypoints.
\newblock In \emph{ECCV}, 2018.

\bibitem[Li et~al.(2021)Li, Si, Li, Hsieh, and Bengio]{li2021learnable}
Yang Li, Si Si, Gang Li, Cho-Jui Hsieh, and Samy Bengio.
\newblock Learnable fourier features for multi-dimensional spatial positional encoding.
\newblock In \emph{NeurIPS}, 2021.

\bibitem[Lin et~al.(2017{\natexlab{a}})Lin, Doll{\'a}r, Girshick, He, Hariharan, and Belongie]{lin2017feature}
Tsung-Yi Lin, Piotr Doll{\'a}r, Ross Girshick, Kaiming He, Bharath Hariharan, and Serge Belongie.
\newblock Feature pyramid networks for object detection.
\newblock In \emph{CVPR}, 2017{\natexlab{a}}.

\bibitem[Lin et~al.(2017{\natexlab{b}})Lin, Goyal, Girshick, He, and Doll{\'a}r]{lin2017focal}
Tsung-Yi Lin, Priya Goyal, Ross Girshick, Kaiming He, and Piotr Doll{\'a}r.
\newblock Focal loss for dense object detection.
\newblock In \emph{ICCV}, 2017{\natexlab{b}}.

\bibitem[Liu et~al.(2021)Liu, Lin, Cao, Hu, Wei, Zhang, Lin, and Guo]{liu2021swin}
Ze Liu, Yutong Lin, Yue Cao, Han Hu, Yixuan Wei, Zheng Zhang, Stephen Lin, and Baining Guo.
\newblock Swin transformer: Hierarchical vision transformer using shifted windows.
\newblock In \emph{ICCV}, 2021.

\bibitem[Loshchilov and Hutter(2019)]{loshchilov2017decoupled}
Ilya Loshchilov and Frank Hutter.
\newblock Decoupled weight decay regularization.
\newblock In \emph{ICLR}, 2019.

\bibitem[Mack and Eckstein(2011)]{mack2011object}
Stephen~C Mack and Miguel~P Eckstein.
\newblock Object co-occurrence serves as a contextual cue to guide and facilitate visual search in a natural viewing environment.
\newblock \emph{Journal of vision}, 11\penalty0 (9):\penalty0 9--9, 2011.

\bibitem[Masciocchi et~al.(2009)Masciocchi, Mihalas, Parkhurst, and Niebur]{masciocchi2009everyone}
Christopher~Michael Masciocchi, Stefan Mihalas, Derrick Parkhurst, and Ernst Niebur.
\newblock Everyone knows what is interesting: Salient locations which should be fixated.
\newblock \emph{Journal of vision}, 9\penalty0 (11):\penalty0 25--25, 2009.

\bibitem[Mondal et~al.(2023)Mondal, Yang, Ahn, Samaras, Zelinsky, and Hoai]{mondal2023gazeformer}
Sounak Mondal, Zhibo Yang, Seoyoung Ahn, Dimitris Samaras, Gregory Zelinsky, and Minh Hoai.
\newblock Gazeformer: Scalable, effective and fast prediction of goal-directed human attention.
\newblock In \emph{CVPR}, 2023.

\bibitem[Najemnik and Geisler(2005)]{najemnik2005optimal}
Jiri Najemnik and Wilson~S Geisler.
\newblock Optimal eye movement strategies in visual search.
\newblock \emph{Nature}, 434\penalty0 (7031):\penalty0 387--391, 2005.

\bibitem[Needleman and Wunsch(1970)]{needleman1970general}
Saul~B Needleman and Christian~D Wunsch.
\newblock A general method applicable to the search for similarities in the amino acid sequence of two proteins.
\newblock \emph{Journal of molecular biology}, 48\penalty0 (3):\penalty0 443--453, 1970.

\bibitem[Oberauer(2019)]{oberauer2019working}
Klaus Oberauer.
\newblock Working memory and attention—a conceptual analysis and review.
\newblock \emph{Journal of cognition}, 2019.

\bibitem[Olivers and Roelfsema(2020)]{olivers2020attention}
Christian~NL Olivers and Pieter~R Roelfsema.
\newblock Attention for action in visual working memory.
\newblock \emph{Cortex}, 131:\penalty0 179--194, 2020.

\bibitem[Park et~al.(2019)Park, Bhattacharya, Yang, Dasari, Das, and Samaras]{park2019advancing}
Sohee Park, Arani Bhattacharya, Zhibo Yang, Mallesham Dasari, Samir~R Das, and Dimitris Samaras.
\newblock Advancing user quality of experience in 360-degree video streaming.
\newblock In \emph{IFIP Networking}, 2019.

\bibitem[Park et~al.(2021)Park, Bhattacharya, Yang, Das, and Samaras]{park2021mosaic}
Sohee Park, Arani Bhattacharya, Zhibo Yang, Samir~R Das, and Dimitris Samaras.
\newblock Mosaic: Advancing user quality of experience in 360-degree video streaming with machine learning.
\newblock \emph{IEEE Transactions on Network and Service Management}, 18\penalty0 (1):\penalty0 1000--1015, 2021.

\bibitem[Ranftl et~al.(2021)Ranftl, Bochkovskiy, and Koltun]{ranftl2021vision}
Ren{\'e} Ranftl, Alexey Bochkovskiy, and Vladlen Koltun.
\newblock Vision transformers for dense prediction.
\newblock In \emph{ICCV}, 2021.

\bibitem[Rashidi et~al.(2020)Rashidi, Ehinger, Turpin, and Kulik]{rashidi2020optimal}
Shima Rashidi, Krista Ehinger, Andrew Turpin, and Lars Kulik.
\newblock Optimal visual search based on a model of target detectability in natural images.
\newblock In \emph{NeurIPS}, 2020.

\bibitem[Sun et~al.(2019)Sun, Chen, and Wu]{sun2019visual}
Wanjie Sun, Zhenzhong Chen, and Feng Wu.
\newblock Visual scanpath prediction using ior-roi recurrent mixture density network.
\newblock \emph{IEEE transactions on pattern analysis and machine intelligence}, 43\penalty0 (6):\penalty0 2101--2118, 2019.

\bibitem[Torralba et~al.(2006)Torralba, Oliva, Castelhano, and Henderson]{torralba2006contextual}
Antonio Torralba, Aude Oliva, Monica~S Castelhano, and John~M Henderson.
\newblock Contextual guidance of eye movements and attention in real-world scenes: the role of global features in object search.
\newblock \emph{Psychological review}, 113\penalty0 (4):\penalty0 766, 2006.

\bibitem[Touvron et~al.(2021)Touvron, Cord, Douze, Massa, Sablayrolles, and J{\'e}gou]{touvron2021training}
Hugo Touvron, Matthieu Cord, Matthijs Douze, Francisco Massa, Alexandre Sablayrolles, and Herv{\'e} J{\'e}gou.
\newblock Training data-efficient image transformers \& distillation through attention.
\newblock In \emph{ICML}, 2021.

\bibitem[Vaswani et~al.(2017)Vaswani, Shazeer, Parmar, Uszkoreit, Jones, Gomez, Kaiser, and Polosukhin]{vaswani2017attention}
Ashish Vaswani, Noam Shazeer, Niki Parmar, Jakob Uszkoreit, Llion Jones, Aidan~N Gomez, {\L}ukasz Kaiser, and Illia Polosukhin.
\newblock Attention is all you need.
\newblock In \emph{NeurIPS}, 2017.

\bibitem[Wang and Shen(2017)]{wang2017deep}
Wenguan Wang and Jianbing Shen.
\newblock Deep visual attention prediction.
\newblock \emph{IEEE Transactions on Image Processing}, 27\penalty0 (5):\penalty0 2368--2378, 2017.

\bibitem[Wang et~al.(2019)Wang, Shen, Xie, Cheng, Ling, and Borji]{wang2019revisiting}
Wenguan Wang, Jianbing Shen, Jianwen Xie, Ming-Ming Cheng, Haibin Ling, and Ali Borji.
\newblock Revisiting video saliency prediction in the deep learning era.
\newblock \emph{IEEE transactions on pattern analysis and machine intelligence}, 43\penalty0 (1):\penalty0 220--237, 2019.

\bibitem[Wang and Klein(2010)]{wang2010searching}
Zhiguo Wang and Raymond~M Klein.
\newblock Searching for inhibition of return in visual search: A review.
\newblock \emph{Vision research}, 50\penalty0 (2):\penalty0 220--228, 2010.

\bibitem[Wolfe(1998)]{wolfe1998visual}
JM Wolfe.
\newblock Visual search. pashler, h.(ed.), attention, 1998.

\bibitem[Wolfe and Horowitz(2017)]{wolfe2017five}
Jeremy~M Wolfe and Todd~S Horowitz.
\newblock Five factors that guide attention in visual search.
\newblock \emph{Nature Human Behaviour}, 1\penalty0 (3):\penalty0 1--8, 2017.

\bibitem[Xie et~al.(2021)Xie, Wang, Yu, Anandkumar, Alvarez, and Luo]{xie2021segformer}
Enze Xie, Wenhai Wang, Zhiding Yu, Anima Anandkumar, Jose~M Alvarez, and Ping Luo.
\newblock Segformer: Simple and efficient design for semantic segmentation with transformers.
\newblock In \emph{NeurIPS}, 2021.

\bibitem[Xu et~al.(2014)Xu, Jiang, Wang, Kankanhalli, and Zhao]{xu2014predicting}
Juan Xu, Ming Jiang, Shuo Wang, Mohan~S Kankanhalli, and Qi Zhao.
\newblock Predicting human gaze beyond pixels.
\newblock \emph{Journal of vision}, 14\penalty0 (1):\penalty0 28--28, 2014.

\bibitem[Yang et~al.(2020)Yang, Huang, Chen, Wei, Ahn, Zelinsky, Samaras, and Hoai]{yang2020predicting}
Zhibo Yang, Lihan Huang, Yupei Chen, Zijun Wei, Seoyoung Ahn, Gregory Zelinsky, Dimitris Samaras, and Minh Hoai.
\newblock Predicting goal-directed human attention using inverse reinforcement learning.
\newblock In \emph{CVPR}, 2020.

\bibitem[Yang et~al.(2022)Yang, Mondal, Ahn, Zelinsky, Hoai, and Samaras]{yang2022target}
Zhibo Yang, Sounak Mondal, Seoyoung Ahn, Gregory Zelinsky, Minh Hoai, and Dimitris Samaras.
\newblock Target-absent human attention.
\newblock In \emph{ECCV}, 2022.

\bibitem[Yarbus(1967)]{yarbus1967eye}
AL Yarbus.
\newblock Eye movements and vision plenum.
\newblock \emph{New York}, 1967.

\bibitem[Zelinsky(2008)]{zelinsky2008theory}
Gregory Zelinsky.
\newblock A theory of eye movements during target acquisition.
\newblock \emph{Psychological review}, 115\penalty0 (4):\penalty0 787, 2008.

\bibitem[Zelinsky et~al.(2019)Zelinsky, Yang, Huang, Chen, Ahn, Wei, Adeli, Samaras, and Hoai]{zelinsky2019benchmarking}
Gregory Zelinsky, Zhibo Yang, Lihan Huang, Yupei Chen, Seoyoung Ahn, Zijun Wei, Hossein Adeli, Dimitris Samaras, and Minh Hoai.
\newblock Benchmarking gaze prediction for categorical visual search.
\newblock In \emph{CVPR Workshops}, 2019.

\bibitem[Zelinsky et~al.(2021)Zelinsky, Chen, Ahn, Adeli, Yang, Huang, Samaras, and Hoai]{zelinsky2021predicting}
Gregory~J. Zelinsky, Yupei Chen, Seoyoung Ahn, Hossein Adeli, Zhibo Yang, Lihan Huang, Dimitrios Samaras, and Minh Hoai.
\newblock Predicting goal-directed attention control using inverse-reinforcement learning.
\newblock \emph{Neurons, Behavior, Data analysis, and Theory}, 5\penalty0 (2):\penalty0 1--9, 2021.

\bibitem[Zhang et~al.(2018)Zhang, Feng, Ma, Lim, Zhao, and Kreiman]{zhang2018finding}
Mengmi Zhang, Jiashi Feng, Keng~Teck Ma, Joo~Hwee Lim, Qi Zhao, and Gabriel Kreiman.
\newblock Finding any waldo with zero-shot invariant and efficient visual search.
\newblock \emph{Nature communications}, 9\penalty0 (1):\penalty0 1--15, 2018.

\bibitem[Zhu et~al.(2021)Zhu, Su, Lu, Li, Wang, and Dai]{zhu2020deformable}
Xizhou Zhu, Weijie Su, Lewei Lu, Bin Li, Xiaogang Wang, and Jifeng Dai.
\newblock Deformable detr: Deformable transformers for end-to-end object detection.
\newblock In \emph{ICLR}, 2021.

\end{thebibliography}
}

\clearpage
\setcounter{page}{1}
\maketitlesupplementary

\section{Experiments on OSIE and MIT1003}

\setlength{\tabcolsep}{5pt}
\begin{table}[t]
\begin{center}
\begin{tabular}{l|cccc}
\toprule 
 & SS & cIG & cNSS & cAUC \\
\midrule
Human consistency & 0.380 & - & -\\\midrule
Chen \etal~\cite{chen2021predicting} & 0.326 & -1.526 & 2.288 & 0.920\\
HAT & {\bf 0.386} & {\bf 2.434} & {\bf 4.515} & {\bf 0.973}\\
\bottomrule
\end{tabular}
\caption{{\bf Comparing free-viewing scanpath prediction algorithms on OSIE} (rows) using multiple scanpath metrics (columns). The best results are highlighted in bold.}
\label{tb:osie}
\end{center}
\end{table}

\setlength{\tabcolsep}{5pt}
\begin{table}[t]
\begin{center}
\begin{tabular}{l|cccc}
\toprule 
 & SS & cIG  & cNSS & cAUC \\
\midrule
Human consistency & 0.363 & - & - & -\\\midrule
Chen \etal~\cite{chen2021predicting} & 0.260 & 0.042 & 1.408 & 0.927\\
HAT & {\bf 0.364} & {\bf 1.311} & {\bf 2.966} & {\bf 0.956}\\
\bottomrule
\end{tabular}
\caption{Comparing free-viewing scanpath prediction algorithms (rows) on {\bf MIT1003 training set using 5-fold cross validation} using multiple scanpath metrics (columns). The best results are highlighted in bold.}
\label{tb:mit1003}
\end{center}
\end{table}

To further validate the effectiveness of our proposed HAT in free-viewing scanpath prediction, we compare HAT to the previous state-of-the-art method in free-viewing scanpath prediction,  \citet{chen2021predicting}, using the OSIE dataset \cite{xu2014predicting} and the MIT1003 dataset \cite{judd2009learning}. Here we only report SS, cIG, cNSS and cAUC and do not use SemSS because free-viewing attention is bottom-up and does not rely on semantics. Moreover, OSIE and MIT1003 do not contain pixel-wise segmentation annotation which is required in SemSS. \cref{tb:osie} and \cref{tb:mit1003} consistently show that HAT surpasses \citet{chen2021predicting} in all metrics by a large margin especially in cIG and cNSS on both free-viewing datasets. The results are consistent with our findings in Tab. 3 of the main text---HAT accurately predicts the scanpaths (reflected by SS), with well-calibrated confidence (as evidenced by the high cIG and cNSS). 
Additionally, we compare HAT to the best alternative overall, \citet{chen2021predicting}, by evaluating the models trained using COCO-FreeView on an {\it unseen} dataset MIT1003 in \cref{tb:mit1003_unseen}. The results show that HAT outperforms \citet{chen2021predicting} in all metrics and with significant improvement in cIG, cNSS and cAUC. This suggests that Chen~\etal's model is prone to be overconfident, whereas HAT better calibrates the confidence in predicting free-viewing fixations and thus provides a more robust prediction of human attention with better generalizability to unseen datasets.

\setlength{\tabcolsep}{5pt}
\begin{table}[t]
\begin{center}
\begin{tabular}{l|cccc}
\toprule
 & SS & cIG  & cNSS & cAUC \\
\midrule
Human consistency & 0.363 & - & - & -\\\midrule
Chen \etal~\cite{chen2021predicting} & 0.210	&-9.735	&0.186	&0.750\\
HAT & {\bf 0.251}	& {\bf 1.052}	& {\bf 2.577} & {\bf 0.951}\\
\bottomrule
\end{tabular}
\caption{{\bf Generalization to an unseen dataset MIT1003}. Both models are trained on COCO-Freeview. The best results are in bold.}
\label{tb:mit1003_unseen}
\end{center}
\end{table}

\setlength{\tabcolsep}{4pt}
\begin{table}[t]
\begin{center}
\begin{tabular}{l|cc|ccc}
\toprule 
 & \multicolumn{2}{c|}{Target-present} & \multicolumn{2}{c}{Target-absent} \\
 \midrule
Target & Seen & Unseen  & Seen & Unseen \\
\midrule
Bottle & Others & Food & Others & Kitchen\\
Bowl & Others & Kitchen & Others & Kitchen\\
Car & Indoor & Outdoor & Vehicle & Others\\
Chair & Others & Kitchen & Indoor & Outdoor\\
Clock & Others & Building & Others & Office\\
Cup & Others & Office & Food & Others\\
Fork & - & - & Others & Bathroom\\
Keyboard & Office & Others & Others & Bedroom\\
Knife & Food & Others & Others & Bathroom\\
Laptop & Others & Living & Others & Living\\
Microwave & Kitchen & Others & Kitchen & Others\\
Mouse & Office & Others & Others & Office\\
Oven & - & - & Others & Living\\
Potted plant & Indoor & Outdoor & Others & Food\\
Sink & Others & Kitchen & Indoor & Outdoor\\
Stop sign & - & - & Others & Vehicle\\
Toilet & Indoor & Outdoor & Others & Bedroom\\
TV & Others & Office & Indoor & Outdoor\\

\bottomrule
\end{tabular}
\caption{\textbf{Data split for scene-to-scene training and testing.} COCO-Search-18 includes 12 scenes: outdoor, street, building, vehicle, food, eatery, kitchen, bathroom, bedroom, living-room (living), dining-room, office. \textbf{Others} in the table includes some of the scenes excluding the unseen scene. \textbf{Outdoor} in the table includes outdoor, street, building and vehicle. In target-present, fork, oven and stop sign are not splittable because they only contain one scene, so we remove them from testing.}
\label{tb:partition}
\end{center}
\end{table}
\setlength{\tabcolsep}{4pt}
\begin{table}[t]
\begin{center}
\begin{tabular}{c|c|cccc}
\toprule 
 \multicolumn{2}{c|}{} & SS & cIG & cNSS & cAUC \\
\midrule
\multirow{4}{*}{TP} & Human(Seen) & 0.520 & - & -\\
 & HAT(Seen) & 0.499 & 2.074 & 5.032 & 0.976\\
  \cmidrule{2-6}
 & Human(Unseen) & 0.546 & - & -\\
 & HAT(Unseen) & 0.481 & 2.454 & 4.537 & 0.973\\\midrule
 \multirow{4}{*}{TA} & Human(Seen) & 0.364 & - & -\\
 & HAT(Seen) & 0.368 & 1.586 & 2.852 & 0.955\\
 \cmidrule{2-6}
 & Human(Unseen) & 0.416 & - & -\\
 & HAT(Unseen) & 0.416 & 2.075 & 3.115 & 0.962\\
\bottomrule
\end{tabular}
\caption{Quantitative result of scene-to-scene generalization on target-present and target-absent task. The first four columns are analysis of scene-to-scene target-present search, and the last four columns are analysis of scene-to-scene target-absent search. Each search contains human consistency and testing results on seen (first two columns of the search) and unseen scenes (last two columns of the search).}
\label{tb:scene2scene-metrics}
\end{center}
\end{table}

\begin{figure*}[t]
  \centering
    {\includegraphics[width=\linewidth]{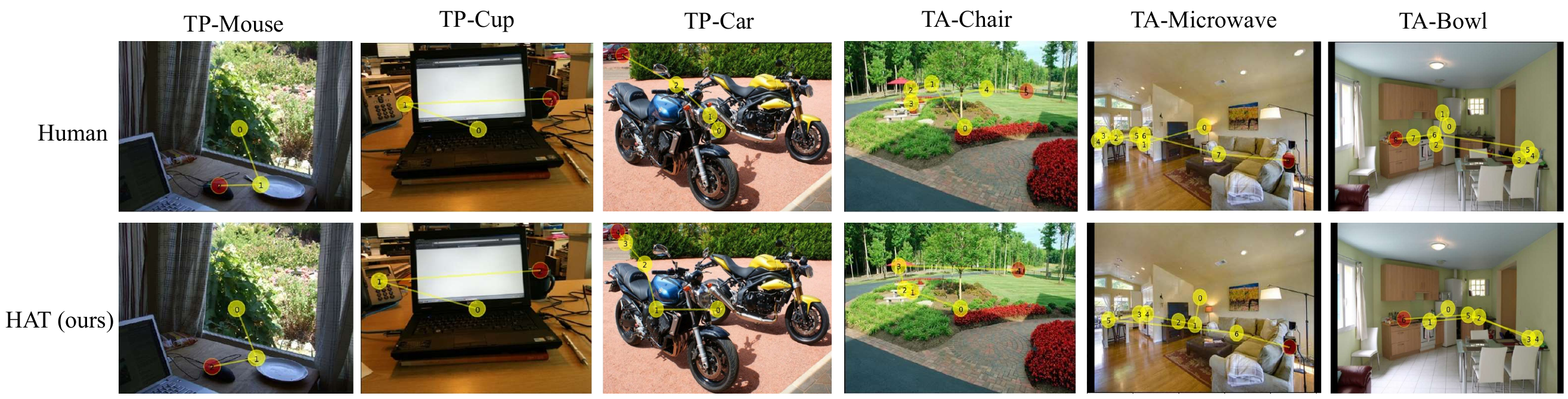}}\label{fig:scene2scene}
    \caption{\textbf{Visual search scanpath visualization for unseen scenes predictions.} The first row is human consistency and the second row is test result on unseen scenes. The first three columns are trained for target-present tasks for mouse, cup and car search, the second three columns are trained for target-absent tasks for chair, microwave and bowl search. }
  \label{fig:scene2scene}
\end{figure*}econd ro
\section{Scene-to-scene Generalization.} 
To further demonstrate the generalization ability of HAT to unseen scenes, we re-partition the COCO-Search18 dataset \cite{chen2021coco} by {\bf scenes}. The new partition contains two test sets: one test set shares the scenes as the training set and the other test set only contains unseen (new) scenes from the training set. We partition COCO-Search18 for each category independently (shown in \cref{tb:partition}). For instance, for the target-absent microwave search task, the training set only contains kitchen scenes while the unseen test set has a variety of other scenes including living rooms, dining rooms, bedrooms and outdoor scenes. To further ensure the unseen images do not exist in the training set, we remove the unseen images for all tasks from the training set because some tasks share the same image stimuli. In the new partition, the target-present set consists of 2170 training images, 568 testing images of unseen scenes, and 273 testing images of seen scenes. The target-absent set consists of 2299 training images, 378 testing images of unseen scenes, and 282 testing images of seen scenes. 

\cref{tb:scene2scene-metrics} presents results of HAT in predicting the TP and TA search scanpaths under both seen and unseen novel scenes. We use human consistency as the baseline. For both tasks, the gap between human consistency and HAT in seen test set is smaller than that in the unseen test set, which is expected. Importantly, HAT's performance on the unseen scenes is on par with human consistency in TA setting although worse in TP setting. Note that the performance difference between seen and unseen human consistency of TA setting is due to the fact that human consistency on the TA test data with seen scenes in the new partition is low. These results suggest that HAT learns to extrapolate from scene to scene and generalize well on novel scenes in visual search scanpath prediction.
The visualization of predicted scanpaths in \cref{fig:scene2scene} further reinforces our observations. By comparing HAT's predicted scanpaths with the ground-truth human scanpaths for unseen scenes in both TP and TA settings, we unveil HAT's robust generalization. For example, in a TP car search task, HAT trained on {\it indoor} scenes successfully located the car on the top-left corner in an {\it outdoor} scene much like humans do. Similarly, when addressing microwave searches under target-absent conditions—with the training set exclusively comprising kitchen scenes—HAT demonstrates significant generalization. This is evident in its proficient extension of predictive capabilities to living-room scenes, as showcased in the fifth column of \cref{fig:scene2scene}. These findings underscore HAT's consistent and robust generalization across diverse scenes, emphasizing its reliable performance in a spectrum of visual search scenarios.

\section{Individual Scanpath Recall}
The importance of predicting personalized scanpath lies in the fact that each person’s unique life experiences shape their individual mental representations of scenes, resulting in personalized perceptions. Therefore, testing the model's ability to generate diversified scanpaths is crucial to learn individual perceptions and to avoid potential biases. To this end, we compute {\it scanpath recall} to measure the extent of individual representation within the model's predictions. For a human scanpath in the stimuli, we consider it to be covered if its sequence score with at least one prediction is higher than the threshold $\tau$. The ratio of covered human scanpaths to all human scanpaths is the recall of the stimuli. \cref{tb:recall} presents the average recall and sequence score of HAT and Gazeformer in both target-present and target-absent search scanpath prediction. For each visual stimuli, we sample 10 scanpaths from the fixation density map and set $\tau$ to the human consistency of sequence score (i.e., $\tau=0.5$ for TP and $\tau=0.381$ for TA). We can see that HAT outperforms Gazeformer in both recall and sequence score by a large margin in target-absent scanpath prediction. This is consistent with our findings in Sec. 4.1 of the main text. Although Gazeformer has a slightly higher sequence score than HAT in TP setting, HAT outperforms Gazeformer significantly in scanpath recall. This implies that HAT better captures the entire scanpath distribution from multiple subjects whereas Gazeformer tends to overfit to an ``average person", thereby repeatedly sampling similar scanpaths given the same image input. 
\setlength{\tabcolsep}{5pt}
\begin{table}[t]
\begin{center}
\begin{tabular}{l|cc|cc}
\toprule 
\multirow{2}{*}{}
 & \multicolumn{2}{c|}{Target-present} & \multicolumn{2}{c}{Target-absent} \\
\cmidrule{2-5}
 & Recall & SS  & Recall & SS \\
\midrule
Gazeformer~\cite{mondal2023gazeformer} & 0.563 & {\bf 0.489} & 0.428 & 0.357\\
HAT & {\bf 0.727} & {0.453} & {\bf 0.750} & {\bf 0.381}\\
\bottomrule
\end{tabular}
\caption{Recall and sequence score comparison between Gazeformer and HAT.}
\label{tb:recall}
\end{center}
\end{table}

\section{Additional Ablation Study}
\label{sec:add_ablation}
\setlength{\tabcolsep}{1.5pt}
\begin{table}[t]
\begin{center}
\begin{tabular}{c|c|ccccc}
\toprule 
 Pixel enc. & Pixel dec. & SemSS  & SS & cIG & cNSS & cAUC\\ 
\midrule
R50& MSD & 0.382 & 0.402 & 1.686 & 3.103 & 0.961 \\
R50& FPN & 0.367 & 0.388 & 1.582 & 2.908 & 0.958\\
R101& MSD & 0.372 & 0.397 & 1.598 & 2.998 & 0.961\\
Swin-B & {MSD} & 0.382 & 0.405 & 1.645 & 3.103 & 0.962\\
\bottomrule
\end{tabular}
\caption{{\bf Comparing different pixel encoder and pixel decoder} in HAT. The ablation experiments are done on the target-absent set of COCO-Search18.}
\label{tb:feat_ext}
\end{center}
\end{table}

\begin{table}[t]
\begin{center}
\begin{tabular}{ccc|cccccc}
\toprule
heads & $\alpha$ & $\beta$ & SemSS & SS & cIG & cNSS & cAUC\\
\midrule
4 & 2 & 4 & \textbf{0.382}  & \textbf{0.402} & \textbf{1.686} & \textbf{3.103} & {0.961} \\
8 & 2 & 4 & 0.375  & 0.390 &  1.310 & 2.826 & 0.961\\
4 & 2 & 2 & 0.381 & 0.401 & 1.129 & 2.633 & 0.960\\
4 & 1 & 4 & 0.378 & 0.393 & 1.566 & 3.046 & \textbf{0.962}\\
\bottomrule
\end{tabular}
\caption{Hyperparameters ablation using
COCO-Search18 TA set.}
\label{tb:params}
\end{center}
\end{table}

\begin{table}[t]
\begin{center}
\begin{tabular}{c|cc}
\toprule 
& Target-present & Target-absent \\
\midrule
Dense & \textbf{0.470} & \textbf{0.403}\\
Regression &  0.452 & 0.330\\
\bottomrule
\end{tabular}

\caption{Comparison between HAT's dense prediction paradigm and Gazeformer's regression paradigm on COCO-Search18 using HAT's architecture.}
\label{tb:regression}
\end{center}
\end{table}



\begin{figure*}[t]
  \centering
    \subfloat[Target-absent laptop search]{\includegraphics[width=.49\linewidth]{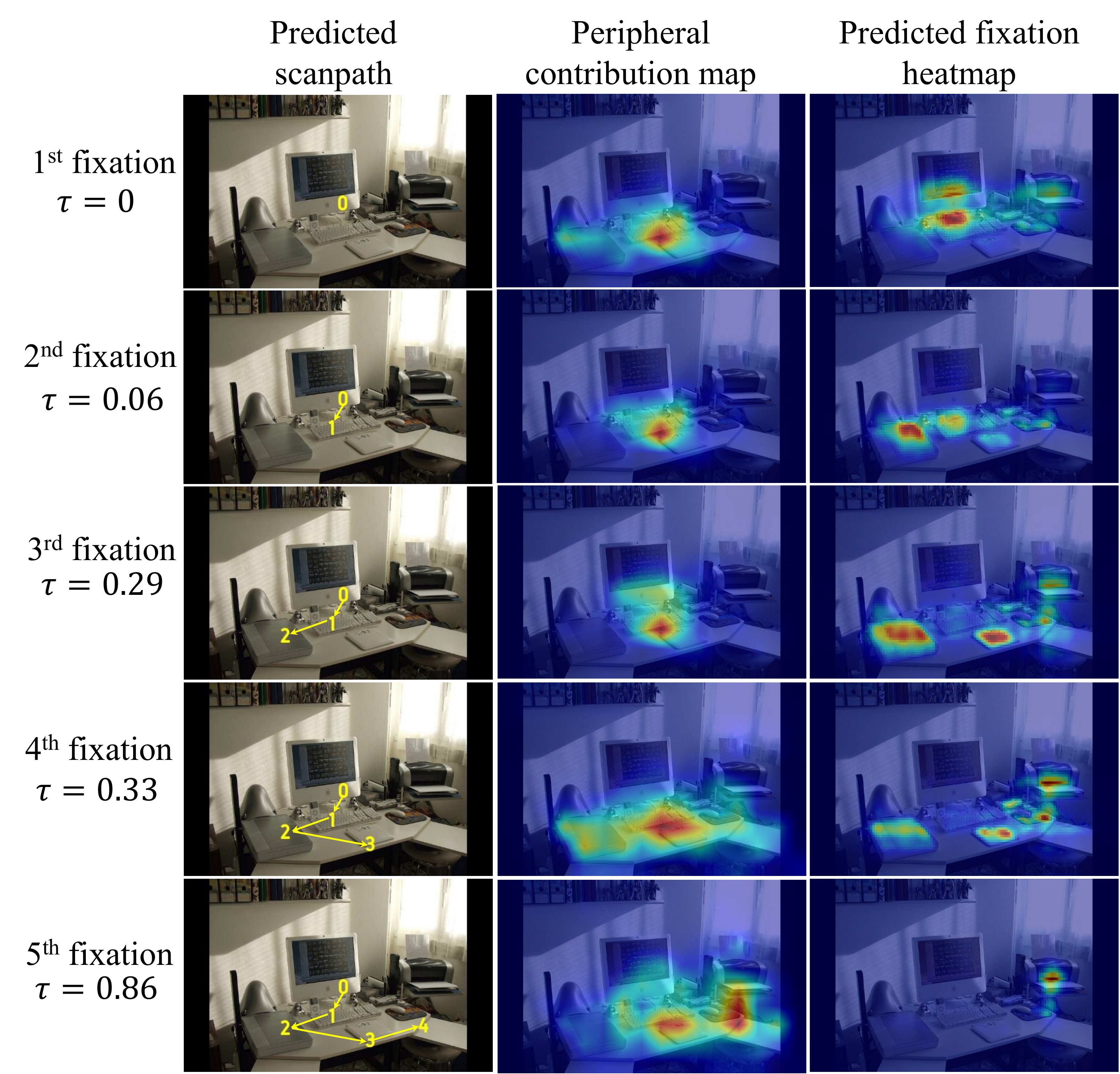}\label{fig:ta_vis2}}
    \subfloat[Target-absent car search]{\includegraphics[width=.49\linewidth]{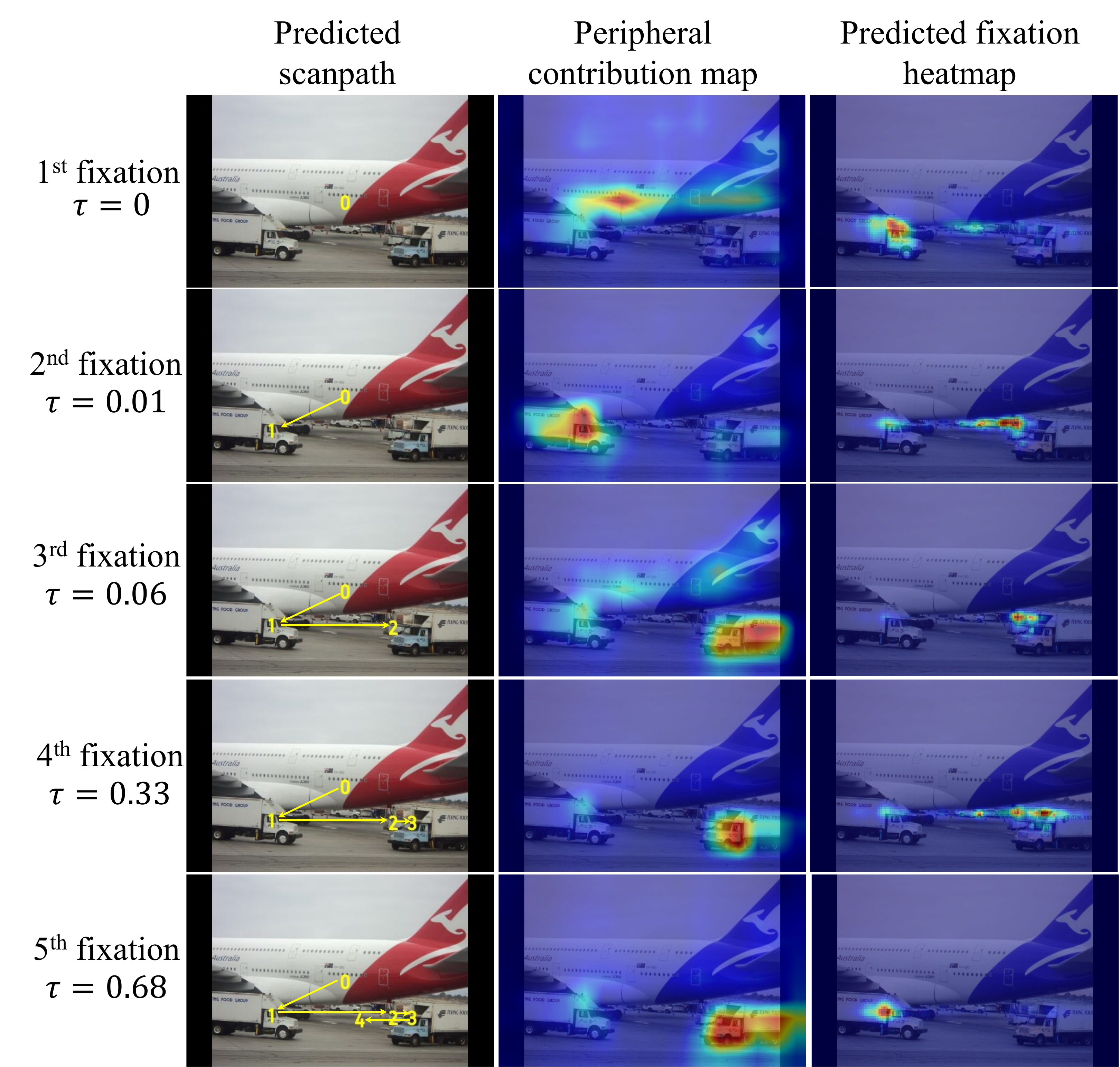}\label{fig:ta_vis1}}
    \caption{Visualization of the {\bf predicted scanpath, peripheral contribution map and fixation heatmap} (columns) of HAT for target-absent (a) laptop and (b) car visual search examples at every fixation (rows). We also include the predicted termination probability $\tau$ for each step on the left. The model terminates searching if $\tau>0.5$.}
    \label{fig:peri_cont}
\end{figure*}

\begin{figure}[t]
  \centering
  \includegraphics[width=1.0\linewidth]{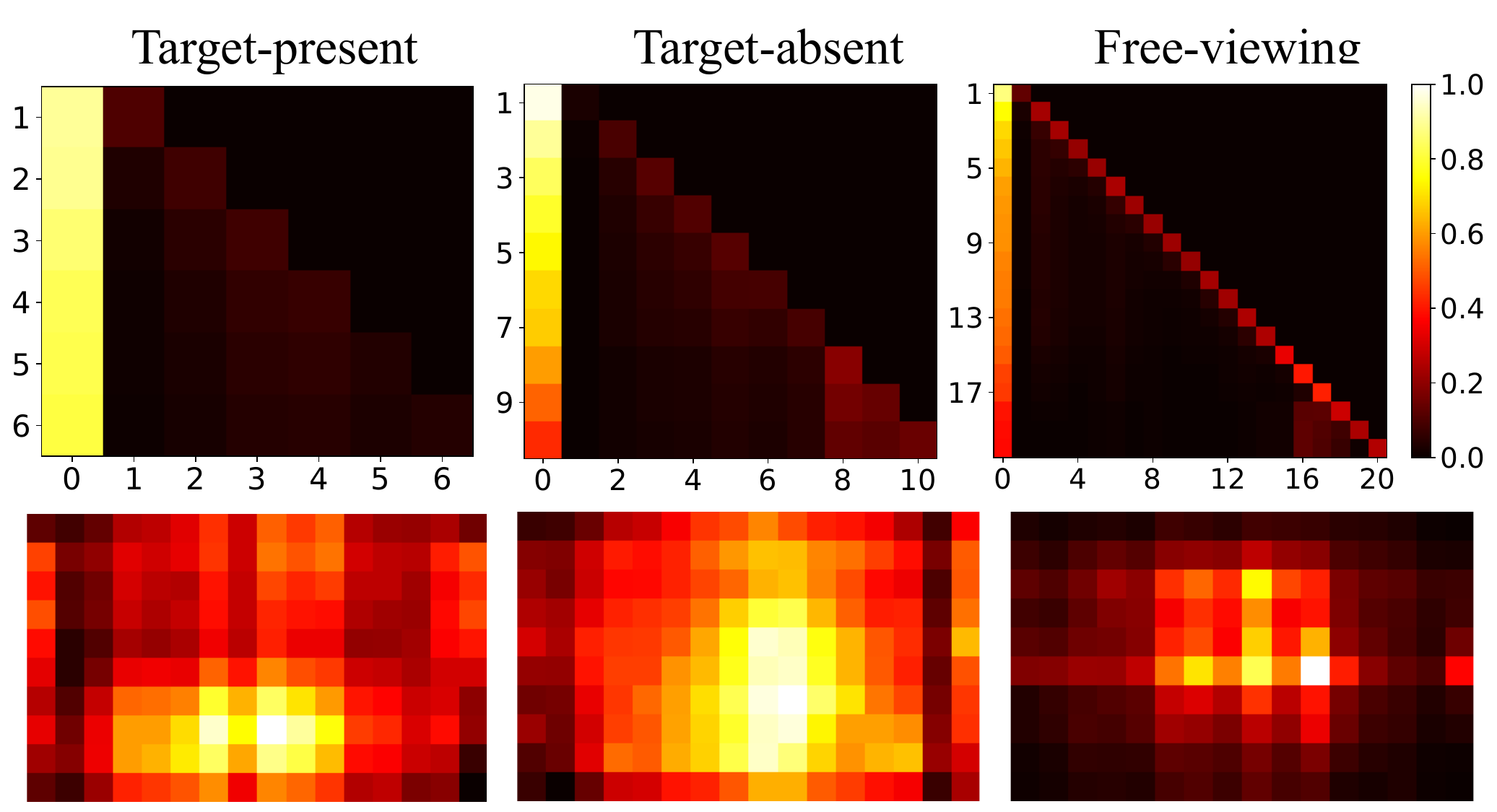}
  \caption{{\bf Peripheral tokens vs foveal tokens} under TP, TA and FV settings (from left to right). 
    The top three figures visualize the temporal change of the contribution of peripheral and foveal memory tokens in predicting human attention. Here the contribution is measured by the attention weight from the last cross-attention layer of the aggregation module in HAT. X-axis shows the token index, with 0 representing all peripheral tokens (by summing the attention weights of all peripheral tokens) and $i>0$ being the $i$-th foveal token. Y-axis indicates temporal fixation step from first to max number of fixation steps allowed for each task. The bottom three figures show the spatial distribution of the attention weights of all peripheral tokens, averaged over the temporal axis. The brighter the color, the larger is the contribution.}
  \label{fig:attn_weights}
\end{figure}

In this section, we provide further ablation on HAT. First we ablate the backbones of HAT. We perform the ablation experiments using the target-absent (TA) visual search fixation prediction task on the TA set of COCO-Search18. By default, HAT uses ResNet-50 \cite{he2016deep} as the pixel encoder and MSD \cite{zhu2020deformable} as the pixel decoder. However, HAT is also compatible with other architectures. Hence, in \cref{tb:feat_ext}, we evaluate HAT with different pixel encoders and decoders. Three pixel encoders: ResNet-50 (R50), ResNet-101 (R101) \cite{he2016deep} and Swin Transformer \cite{liu2021swin} (we use the base model, Swin-B); and two pixel decoders: FPN \cite{lin2017feature} and MSD \cite{zhu2020deformable}, are evaluated. One can observe that MSD is better than FPN as the pixel decoder and HAT performs the best when using R50 and Swin-B as the pixel encoder. Notice that the performance gap between different pixel encoders is small, suggesting that the performance of HAT is robust to the choice of different pixel encoder architectures. 
More importantly, all of these configurations of HAT significantly outperforms all baselines in Tab. 2 of the main text.

In \cref{tb:params}, we also present HAT's results with varied hyperparameters: the number of attention heads in the transformer module of HAT, $\alpha$ and $\beta$ of \eqref{eq:loss_fn}, demonstrating HAT's robustness w.r.t. difference choices of hyperparameters. Notably, the choice of $(4, 2,4)$ in the three ablated hyperparameters achieves the best performance.

\cref{tb:regression} compares HAT's DP task with Gazeformer's Reg task using HAT in TP and TA settings. The proposed DP outperforms Reg in both settings, especially in TA setting. This aligns with our findings in Tab. 1-3 of the main text which show that Gazeformer's Reg paradigm, assuming a Gaussian fixation distribution, is less effective for TA and FV scanpaths.

\section{Additional Qualitative Analysis}

\subsection{Model interpretability}
\myheading{Peripheral contribution map visualization.}
In Sec. 4.2 of the main text, we showed that the peripheral contribution maps in HAT can be leveraged to interpret the model's behaviors using a target-present search example. We also observe a similar pattern in the target-absent (TA) setting (see \cref{fig:peri_cont}). In \cref{fig:ta_vis2}, we see that in a TA laptop search task pixels of {\it table} and {\it keyboard} contribute significantly in predicting human fixations as tables and keyboards can provide spatial cues for the laptop. This reveals a unique factor that guides visual search attention---anchor objects \cite{boettcher2018anchoring}. In \cref{fig:ta_vis1}, we see that TA car search fixations are attracted to truck pixels as trucks and cars are closely related concepts that are considered as distractors.

\myheading{Peripheral vs foveal.}
We also \textit{collectively} analyze the contribution of peripheral tokens and foveal tokens in predicting human attention control under the TP, TA and FV settings, separately. \cref{fig:attn_weights} visualizes the temporal change of contributions of all peripheral tokens collectively and the foveal token in predicting human attention averaged over all test images. We observe that the peripheral tokens contribute the most in predicting TP fixations across all fixations (forming the yellow column on the left). This is because in TP images there is a strong target signal available in the visual periphery to guide attention. Contrast this with FV fixations, where the contribution of the peripheral tokens diminishes over the temporal space and the only the current foveal token has a strong and consistent contribution (a clear red diagonal line). An interpretation of this pattern is that people have only a poor memory of what they viewed in previous fixations and their attention is controlled by salient pixels within a local neighborhood around the current fixation. Interestingly, for TA fixations we also observe a diminishing contribution of the peripheral tokens over the temporal space, but not as pronounced. Moreover, as more fixations are made, the contribution of recent fixations increases, approaching the pattern in FV. This suggests that the later fixations of a TA scanpath behave like a FV scanpath, which confirms a finding in \cite{Chen_2022_CVPR}. Lastly, the bottom row visualizes the contribution of each individual peripheral token (averaged over the temporal axis), where we see peripheral tokens encode a strong center bias for FV fixations, whereas TA fixations show a weaker center bias and TP fixations show no obvious center bias at all, again as expected and confirming previous suggestion. This showcases the potential for HAT to make highly interpretable predictions of human attention control.

\begin{figure}[t]
  \centering
  \includegraphics[width=1.\linewidth]{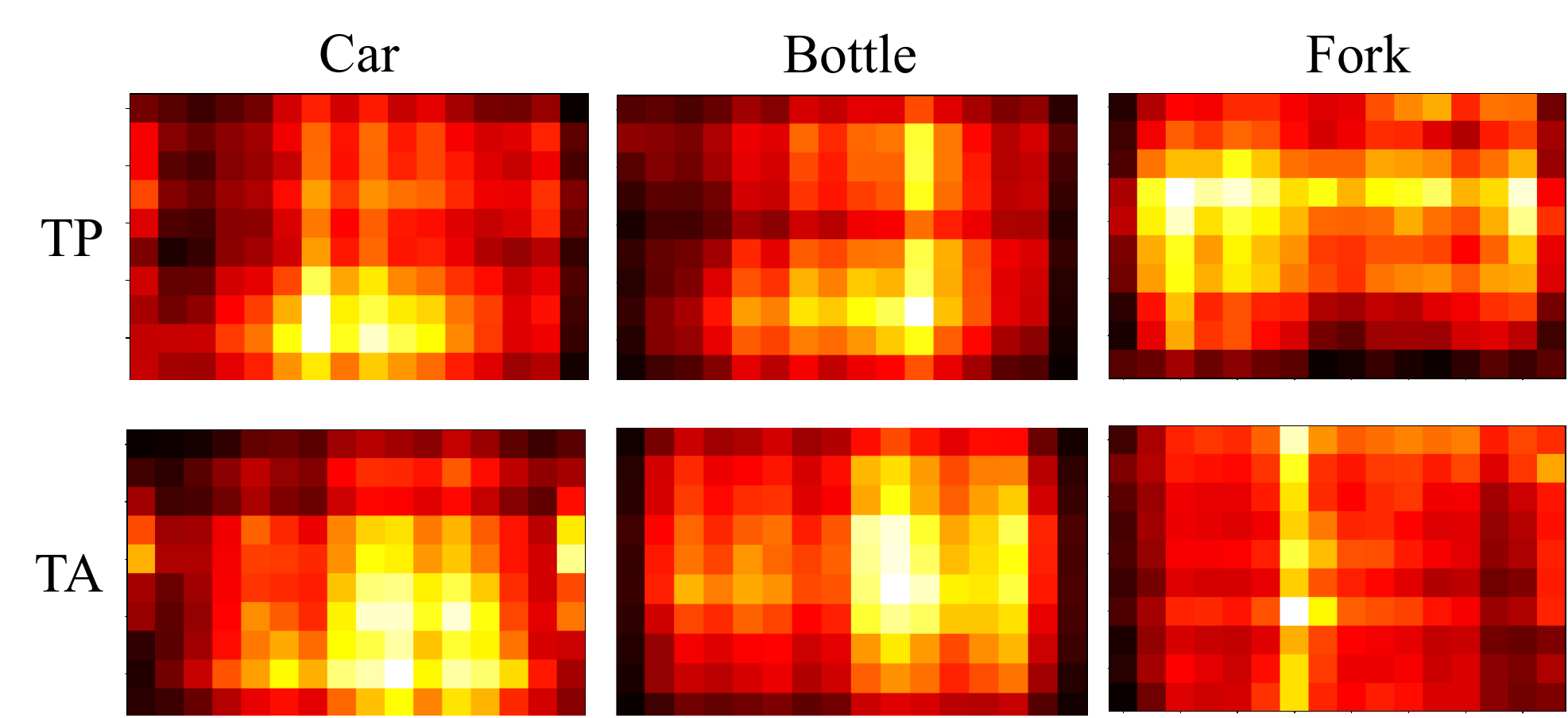}
  \caption{{\bf Categorical peripheral contribution map of visual search fixations}. We show the contribution map of the peripheral tokens for two categories (rows): car and bottle, in target-present and target-absent settings (columns). We measure the contribution of each peripheral token by the attention weights from the last cross-attention layer of the aggregation module in HAT, averaged over the temporal axis of all testing data in COCO-Search18 \cite{chen2021coco}. The brighter the color, the larger the contribution.}
  \label{fig:cat}
\end{figure}
\begin{figure*}[t]
  \centering
{\includegraphics[width=1.0\linewidth]{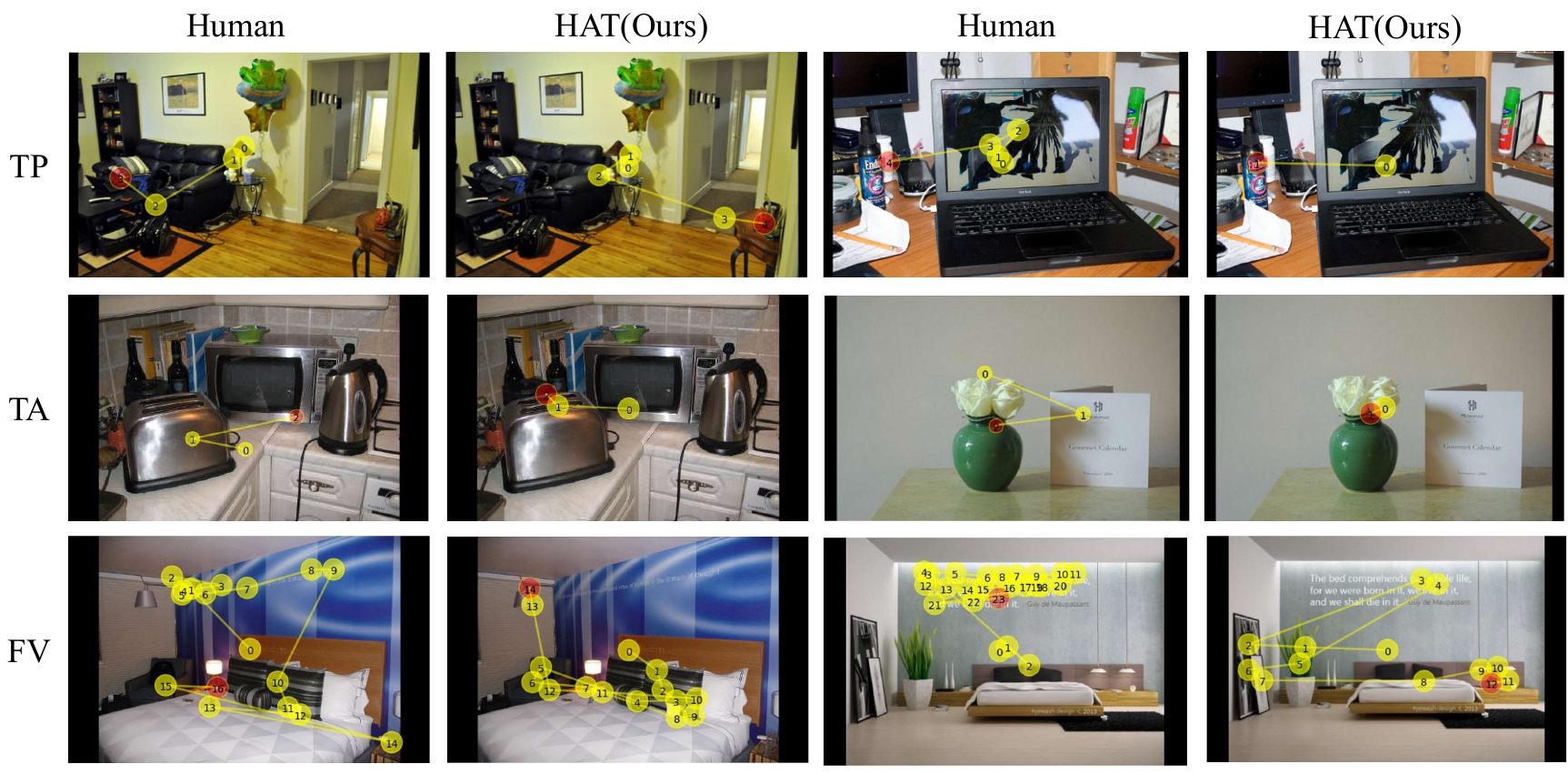}\label{fig:failcase}}
\caption{\textbf{Failure cases.} The first row is two failure cases for laptop and tv search, respectively. The second row is two failure cases for sink and clock search, respectively. The third row is two failure cases for free viewing. }
    \label{fig:failcase}
\end{figure*}
\myheading{Target prior.}
A natural question arising from this observation is whether the peripheral tokens of TA and TP fixations encode a {\it target prior}--spatial distribution of the possible target location. To answer this question, we visualize the category-wise peripheral contribution maps for TP and TA fixations by averaging the attention weights (on the peripheral tokens) of the last cross-attention layer over all testing fixations for each target category. As shown in \cref{fig:cat}, the category-specific peripheral contribution map does not provide a clear evidence of TA and TP peripheral contribution map being a target prior, but we find some target-specific pattern, e.g., the contribution is pronounced around the bottom horizontal area for ``car" and around the vertical area for ``bottle", which may represent the spatial prior of each category. 

\subsection{Failure cases analysis}

Our analysis of failure cases offers insights for future research. A scanpath prediction would be taken as a failure case if its sequence score falls below 50\% of the human consistency of its stimulus. Under this criterion, we find some common features of failure cases. For target-present, the ambiguity of the target object  often leads to a decline in HAT performance. For instance, in the first row of \cref{fig:failcase}, the laptop in the first case has a very similar color to the table and its surrounding objects. In the second case, the TV is indistinguishable even to an individual. Both scenarios present an ambiguous visual representation of the target, complicating the prediction of the scanpath during visual searches. For target-absent, it is hard for HAT to learn the perception pattern of human when the human scanpaths are very short. In free-viewing, from the visualization in the third row of \cref{fig:failcase}, HAT only allocates a few fixations to text in the image, which is opposite to human perception. This discrepancy is attributed to the limitations of the image encoder and decoder in capturing text features.

\begin{figure*}[t]
  \centering
  \includegraphics[width=1.0\linewidth]{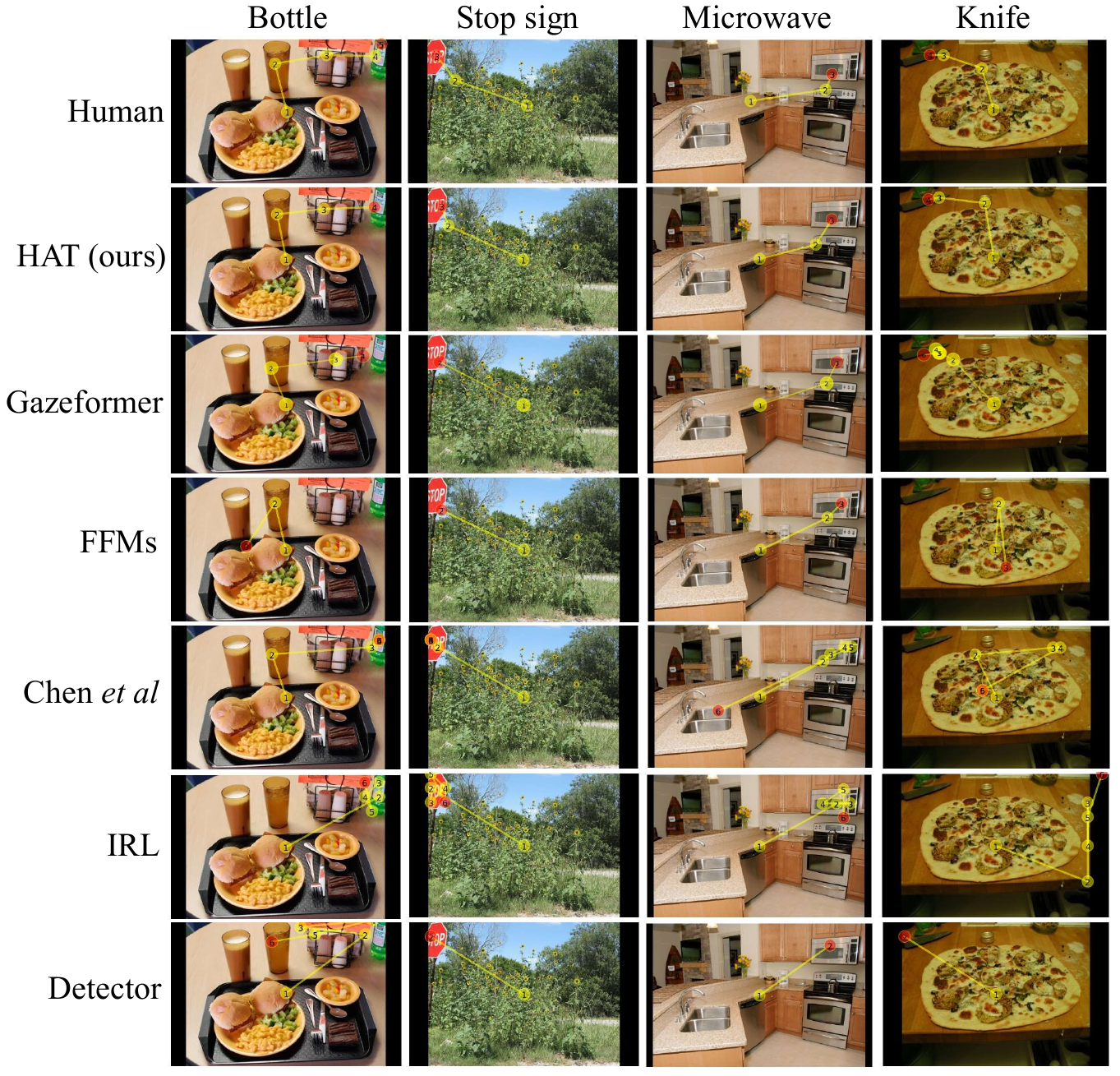}
  \caption{{\bf Target-present scanpath visualization}. We show the scanpaths of seven methods (rows) for four different targets (columns) which are bottle, stop sign, microwave and knife. The final fixation of each scanpath is highlighted in red circle. For methods without termination prediction, i.e., IRL and detector, we visualize the first 6 fixations.}
  \label{fig:sps_tp}
\end{figure*}

\begin{figure*}[t]
  \centering
  \includegraphics[width=1.0\linewidth]{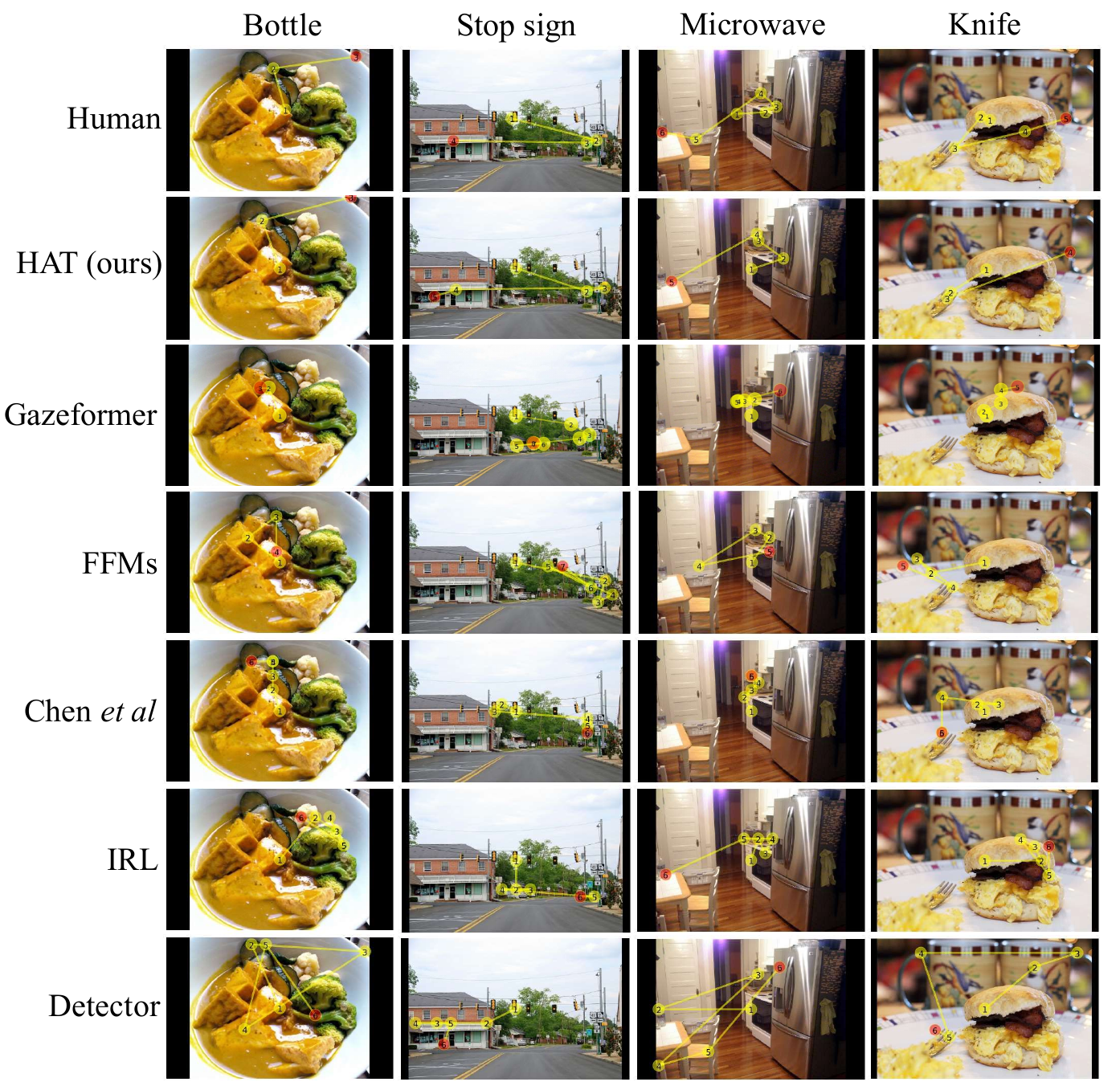}
  \caption{{\bf Target-absent scanpath visualization}. We show the scanpaths of seven methods (rows) for four different targets (columns) which are bottle, stop sign, microwave and knife. The final fixation of each scanpath is highlighted in red circle. For methods without termination prediction, i.e., IRL and detector, we visualize the first 6 fixations.}
  \label{fig:sps_ta}
\end{figure*}

\begin{figure*}[t]
  \centering
  \includegraphics[width=1.0\linewidth]{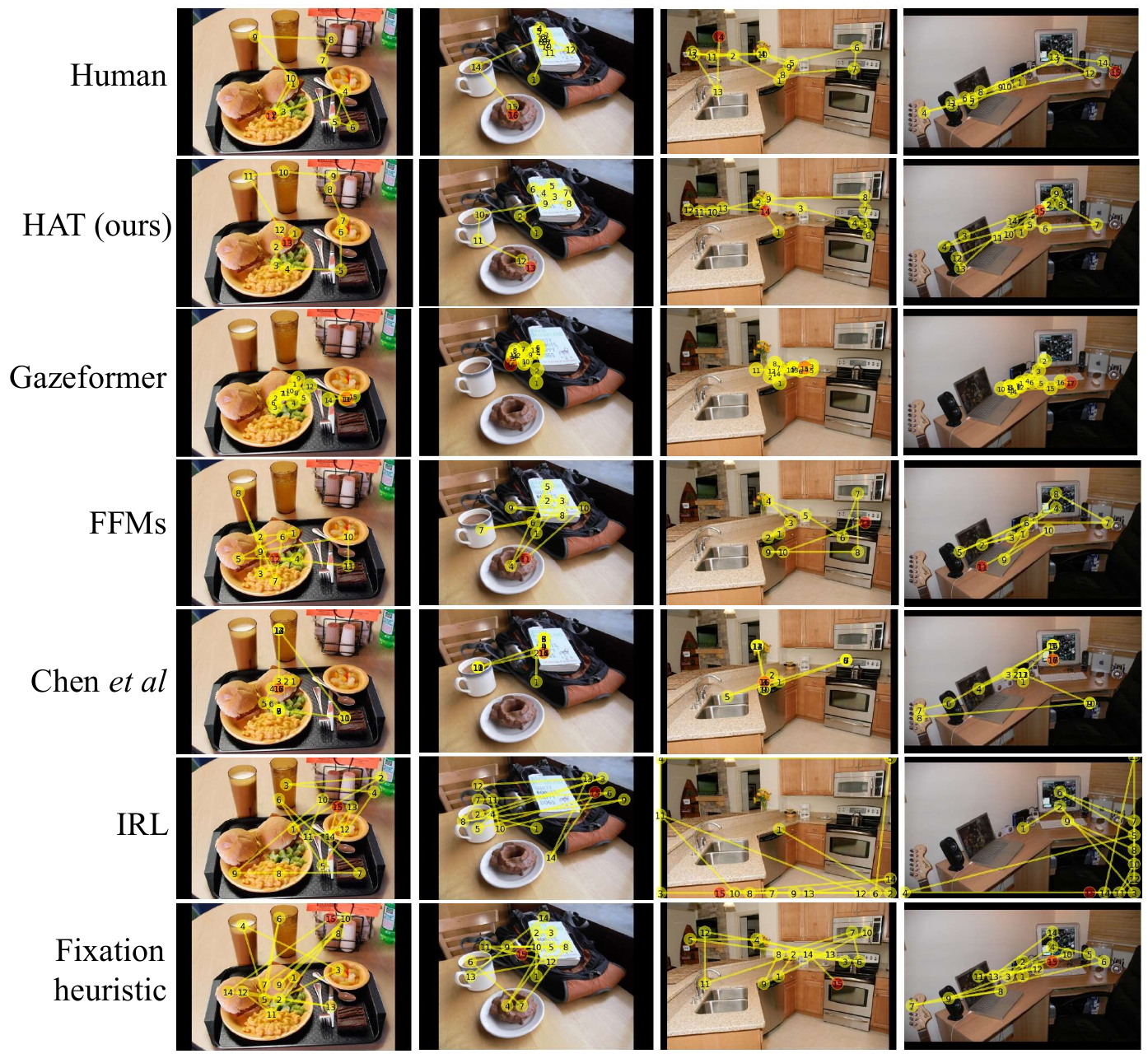}
  \caption{{\bf Free-viewing scanpath visualization}.  We show the scanpaths of seven methods (rows) for four example images. The final fixation of each scanpath is highlighted in red circle. For methods without termination prediction, i.e., IRL and detector, we visualize the first 15 fixations.}
  \label{fig:sps_fv}
\end{figure*}
\subsection{Scanpath visualization}
We further visualize additional scanpaths for human (ground truth), our HAT, Gazeformer \cite{mondal2023gazeformer}, FFMs \cite{yang2022target}, Chen \etal \cite{chen2021predicting}, IRL \cite{yang2020predicting}, and a heuristic method (target detector for visual search and saliency heuristic for free viewing) in the TP, TA, and FV settings. \cref{fig:sps_tp} shows the TP scanpaths. In all examples, HAT shows superior performance in predicting the human fixation trajectory not only when humans correctly fixate on the target, but also when their attention is distracted by other visually similar objects. For example, in the last column of \cref{fig:sps_tp} when the task is to find a knife, HAT is the only model that correctly predicts the fixation on the metallic object (because knives are usually metallic), whereas other methods either missed the target or did not show any distractions to the metallic object. This shows the capacity of HAT in modeling human attention control in visual search. \cref{fig:sps_ta} shows that HAT learns to leverage the context cues in predicting target-absent fixations, e.g., when the search target is microwave, HAT correctly predicted the fixations on the counter-top and table, where microwaves are often found. Similarly, HAT also generates the most human-like scanpaths in free-viewing task (see \cref{fig:sps_fv}), capturing all important aspects of scanpaths, such as the locations (where), the semantics (what), and the order (when) of the fixations.

\section{Implementation details}
\subsection{Network structure.}
HAT has four modules as shown in Fig. 2 of the main text. By default, the feature extraction module employs ResNet-50 \cite{he2016deep} as the pixel encoder and MSDeformAttn \cite{zhu2020deformable} as the pixel decoder. \cref{{sec:add_ablation}} presents the results with other pixel encoders, ResNet-101 and Swin Transformer \cite{liu2021swin}, and pixel decoder, FPN \cite{lin2017feature}.
The number of channels of the feature maps $C$ is set to 256. For the foveation module, the transformer encoder has three layers. The transformer decoder in the aggregation module has six layers (i.e,. $L=6$). All transformer encoder and decoder layers in HAT have 4 attention heads. The number of queries $N=18$ for visual search scanpath prediction as COCO-Search18 contains 18 target categories and $N=1$ for free-viewing scanpath prediction. Finally, the MLP in the fixation prediction module has two linear layers with 512 hidden dimensions and a ReLU activation function.

\subsection{Training settings.}
Following \cite{yang2022target,yang2020predicting}, we resize all images to $320{\times} 512$ for computational efficiency during training and inference. We use the AdamW \cite{loshchilov2017decoupled} with the learning rate of $0.0001$ and train HAT for 30 epochs with a batch size of 32. No data augmentation is used during training. Note that we keep the pixel encoder fixed during training and we use the COCO-pretrained weights for panoptic segmentation from \cite{cheng2022masked} as an initialization for the pixel encoder and pixel decoder.
Following \cite{yang2020predicting}, we set the maximum length of each predicted scanpath to 6 and 10 (excluding the initial fixation) for target-present and target-absent search scanpath prediction, respectively. For free viewing, the maximum scanpath length is set to 20.

\subsection{Additional details on heuristic baselines} 
\textbf{Detector}: The detector network consists of a feature pyramid network (FPN) for feature extraction (1024 channels) with a ResNet50 pretrained on ImageNet as the backbone and two convolution layers with batch normalization and a ReLU activation layer in between for detection of 18 different targets. The kernel size and hidden dimension of the first convolutional layer is 3 and 128, respectively. The detector network predicts a 2D spatial probability map of the target from the image input and is trained using the ground-truth location of the target. Another similar baseline is \textbf{Fixation Heuristic}.
This network shares exactly the same network architecture with the detector baseline but it is trained with behavioral fixations in the form of spatial fixation density map, which is generated from 10 subjects on the training images.

\subsection{Scanpath generation}
Most methods except human consistency generate a new spatial priority map or action map at every step, while the predicted priority map is fixed over all steps for the Detector, Fixation Heuristic and IVSN baselines. Prior works like \cite{yang2020predicting} measure model performance based on multiple randomly sampled scanpaths, which, however, can be unfairly bias toward models that repeatedly sample the best (greedy) scanpath. Therefore, in this work we directly compare different methods using their best predictions. When generating scanpaths, for all methods we follow \cite{yang2022target} and predict one scanpath for each testing image in a greedy fashion---a fixation location is determined by selecting the most probable fixation location in the predicted priority map. At evaluation, we compare the predicted ``greedy" scanpath against all GT scanpaths which helps measure how well a model (at its best) captures human scanpath consistency.

\subsection{Implementation of cIG}
cIG denotes the amount of information gain from the predicted fixation map (the model is provided with all previous fixations) over a baseline in predicting the ground-truth fixation. Here, the baseline is a  fixation density map constructed by averaging the smoothed density (with a Gaussian kernel of one degree of visual angle) maps of all training fixations. For target-present and target-absent visual search settings, we use a (target) category-wise fixation density map, following \cite{yang2022target}. For the heuristic models (i.e., target detector and saliency heuristic) which apply the winner-take-all strategy on a static fixation map to generate the scanpath prediction, we use the same static fixation map for all fixations in a scanpath to compute cIG, cNSS and cAUC. To obtain the predicted fixation maps for Chen \etal's model~\cite{chen2021predicting}, we use the ground-truth fixation map (Gaussian smoothed with a kernel size of 2) as input to obtain the predicted action map for the next fixation (i.e., the predicted fixation map). Note that all predicted fixation maps in computing cIG, cNSS and cAUC, are resized to $320\times512$ for fair comparison.

\section{A note on human consistency}
For the same image, there are multiple ground-truth scanpaths from different human subjects. 
As in \cite{yang2020predicting,yang2022target}, for each image human consistency is computed by averaging the similarities of every pair of human scanpaths. A model's performance, however, is measured by the average similarity of the predicted \enquote{mean} scanpath to every human scanpath. 
Consider scanpaths as 2D points whose similarity can be measured by Euclidean distance. The average pairwise similarity between these points (human consistency) is smaller than the average similarity of these points to their arithmetic mean (model performance).
This explains how it is possible for a good model to exceed the human consistency. Taking a triangle as analogy: the average distance of a point (predicted scanpath) within the triangle to all vertices can be smaller than the average edge length (human consistency).

\section{Further discussion on applications}
\label{sec:app} 
Models that predict top-down attention (TP/TA search fixations), modulated by an external goal, have wide applicability to attention-centric HCI. For example, 
faster attention-based rendering that leverages the prediction of a user's attention as they play a VR/AR game and home robots incorporating search-fixation-prediction models will be better at inferring a user's need (i.e., their search target).
Home robots incorporating search-fixation-prediction models will be better able to infer a users' need (i.e., their search target) and autonomous driving systems can attend to image input like an undistracted driving expert.
Applications of FV attention prediction exist in foveated rendering \cite{kaplanyan2019deepfovea} and online video streaming \cite{park2021mosaic}.

\end{document}